%% file: main.tex
\documentclass{article}
\usepackage{uxbench}

\input{math_commands.tex}

\usepackage{amsmath,amssymb}
\usepackage{booktabs,tabularx,array}
\usepackage{xcolor}
\usepackage{url}
\usepackage{booktabs}
\usepackage{multirow}
\usepackage{graphicx}
\usepackage{tcolorbox}
\usepackage{wrapfig}
\usepackage{enumitem}
\usepackage{colortbl}

\usepackage{xspace}

\usepackage{graphicx,xspace,booktabs,amssymb,pifont}
\newcommand{\cmark}{\ding{51}}  
\newcommand{\xmark}{\ding{55}}  
\newcommand{\pmark}{$\circ$}
\definecolor{gaincolor}{HTML}{1B7837}
\newcommand{\gain}[1]{\,\textcolor{gaincolor}{\scriptsize (#1)}}

\usepackage{graphicx,xspace}

\newcommand{\bench}{\textsc{GroupAssistBench}\xspace}

\tcbuselibrary{skins}              
\definecolor{findbg}{HTML}{EEF3FA}
\definecolor{findframe}{HTML}{5B8FD6}
\newtcolorbox{finding}[1][]{%
  enhanced,
  colback=findbg, colframe=findframe,
  boxrule=0pt, leftrule=2pt, arc=1.5mm,
  left=6pt, right=6pt, top=4pt, bottom=4pt,
  before skip=6pt plus 2pt, after skip=8pt plus 2pt,
  fontupper=\small,
  #1}
\newcommand{\findlabel}[1]{\textbf{\textcolor{findframe}{#1}}\quad}

\definecolor{gaincolor}{HTML}{1B7837}
\definecolor{losscolor}{HTML}{B03A2E}
\definecolor{fullcontextbg}{HTML}{F2EFFE}
\definecolor{retrievalbg}{HTML}{D7E9FF}
\definecolor{externalmemorybg}{HTML}{DDF8E9}

\title{One Assistant, Many Memories: Benchmarking Real-World Group Assistance When Recall Is Not Enough}

\author{%
  {\normalsize\bfseries
    Zhanghao Hu$^{1,2,*}$,
    Linhai Zhang$^{2,*}$,
    Qian Zhao$^{1}$,
    Qi Zhu$^{1}$,
    Yuan Hua$^{1}$,\\
    Di Liang$^{1}$,
    Jiasheng Si$^{3}$,
    Xin Zhao$^{4}$,
    Yulan He$^{2}$,
    Zhumin Chen$^{5,\dagger}$,
    Lin Gui$^{2,\dagger}$}\\[0.35em]
  {\footnotesize
    $^{1}$Tencent Yuanbao Team
    \quad $^{2}$King's College London}\\
  {\footnotesize
    $^{3}$Qilu University of Technology
    \quad $^{4}$Renmin University of China
    \quad $^{5}$Shandong University}
}

\begin{document}
\maketitle

\begingroup
\renewcommand{\thefootnote}{\fnsymbol{footnote}}
\footnotetext[0]{$^{*}$ Equal contribution. \qquad $^{\dagger}$ Corresponding authors.}
\endgroup

\begin{abstract}

The value of memory lies not only in recall but in using it to provide appropriate assistance.
Recent benchmarks test memory use in personalisation and agentic tasks, but seldom evaluate assistance without fixed answers or integration across multiple participants.
Group chats make both challenges more acute: an assistant must integrate information from multiple members, track revisions and conflicts over time, and produce one response for the group.
Yet existing group memory benchmarks rely on synthetic conversations and inherit constrained evaluation from general memory benchmarks, focusing on recall against fixed answers or decisions over a closed set.
We introduce \bench, the first memory benchmark from real-world group conversations with a deployed group assistant, covering 80,299 messages in 50 groups and costing over \$53K.\footnote{See Appendix~\ref{app:cost-accounting} for the cost breakdown.}
Over the same histories, \bench evaluates memory recall (L1) with reference answers and memory use (L2) with service requests without fixed answers across single, distributed, and conflicting memories (M1--M3).
Requests are verified to require group history, memories are anchored to their sources, and L2 uses tailored rubrics with penalties derived from real user feedback.
Across evaluations of frontier models with full context, retrieval methods, and memory systems, the strongest memory system rivals full context on recall but falls far behind on use, especially for distributed and conflicting memories.
Together, two additions that relate retrieved memories and bind remembered facts as response constraints bring consistent gains, showing that group memory systems must improve use, not recall alone.

\end{abstract}

\input{sections/1_introduction_v3}

\input{sections/2_related_work_v4}

\input{sections/3_methodology_v3}

\input{sections/4_experiment}

\input{sections/5_analysis_v2}

\input{sections/6_conclusion}

\clearpage

\section*{AI use statement}
In this work, we used generative AI tools to assist with synthetic data generation, dataset cleaning and reformatting, and the design and refinement of research methodology and experiments. We also used generative AI tools to create or modify scientific figures and to summarise and analyse existing literature. Generative AI tools were not used to formulate mathematical claims.  
All AI-assisted outputs were reviewed by the authors. For example, research ideas suggested by LLMs were checked against the existing literature through a manual literature review to guard against unoriginal or unsupported claims, and LLM-generated code was manually reviewed, tested, and validated for correctness. We take full responsibility for the final content of this work, including all text, claims, code, figures, and other artefacts produced with the assistance of generative AI.

\section*{Ethics Statement}
This study uses pseudonymized group-chat records. 
The use of these records for research was reviewed and approved through the data provider’s privacy review process before the data were made available to the authors. 
Speaker identifiers are replaced with group-specific aliases, and message texts are rewritten to remove residual personal information, with safety-filtered content explicitly marked. 
The original source records remain restricted to the data provider and will not be included in any public release. 
Any release of derived benchmark artefacts will undergo an additional privacy and compliance review to further reduce disclosure and re-identification risks.

\section*{Reproducibility statement}
Upon acceptance, we will publicly release all author-owned code and privacy-reviewed derived benchmark artefacts necessary to reproduce the reported results. The benchmark release will include the rewritten and pseudonymized Chinese and English group histories, member profiles, queries, per-instance history cutoffs and split manifests, source-linked anchors and evidence spans, L1 reference and distractor atoms, and L2 rubrics with criterion types and weights. The code release will include the exact task and evaluator prompts, benchmark construction and validation pipeline, evaluator parsing and deterministic scoring code, result aggregation scripts, author-written baseline adapters, configurations, the relational evidence organisation and assistance constraint binding implementations, end-to-end run scripts, dependency specifications, and model and API version information. We will also release the model responses, evaluator judgments, and aggregate run metadata underlying the reported tables so that the reported scores can be verified without rerunning proprietary models. For third-party implementations or model weights that we cannot redistribute, we will provide versioned references together with our patches, adapters, and configurations. All released artefacts will be accompanied by explicit licenses or data-use terms. Original source records and the other restricted materials listed in the Ethics Statement will remain private.

\clearpage

\bibliography{references}
\bibliographystyle{uxbench}

\clearpage
\appendix

\input{sections/appendix_v2}

\end{document}

%% file: math_commands.tex
\usepackage{amsmath,amsfonts,bm}

\def\eqref#1{equation~\ref{#1}}

\def\1{\bm{1}}

\DeclareMathAlphabet{\mathsfit}{\encodingdefault}{\sfdefault}{m}{sl}
\SetMathAlphabet{\mathsfit}{bold}{\encodingdefault}{\sfdefault}{bx}{n}



%% file: sections/1_introduction_v3.tex
\section{Introduction}

Memory is a core component of agents built on LLMs and has been widely studied in recent years~\citep{maharana-etal-2024-evaluating,wu2025longmemeval,DBLP:journals/corr/abs-2512-13564}. 
Recent benchmarks have moved beyond factual recall to evaluate whether remembered information supports personalization~\citep{DBLP:journals/corr/abs-2512-06688,xiao2026alpsbench,DBLP:journals/corr/abs-2606-22877} and agentic task completion~\citep{he2026benchmarking,zhao2026amabench}. 
This shift from remembering information to using it effectively becomes especially challenging in multi-user settings. Group chats provide a particularly demanding setting for this shift from recall to use: unlike a personal assistant serving one user, an assistant in a group chat serves the group as a whole and must produce a single response that reconciles information distributed across members, revised over time, or contested within the group~\citep{DBLP:journals/corr/abs-2605-14498,10.1145/3770855.3817589,DBLP:journals/corr/abs-2605-17789,DBLP:journals/corr/abs-2606-18829}.

\begin{figure}[h]
\centering
\includegraphics[width=\linewidth]{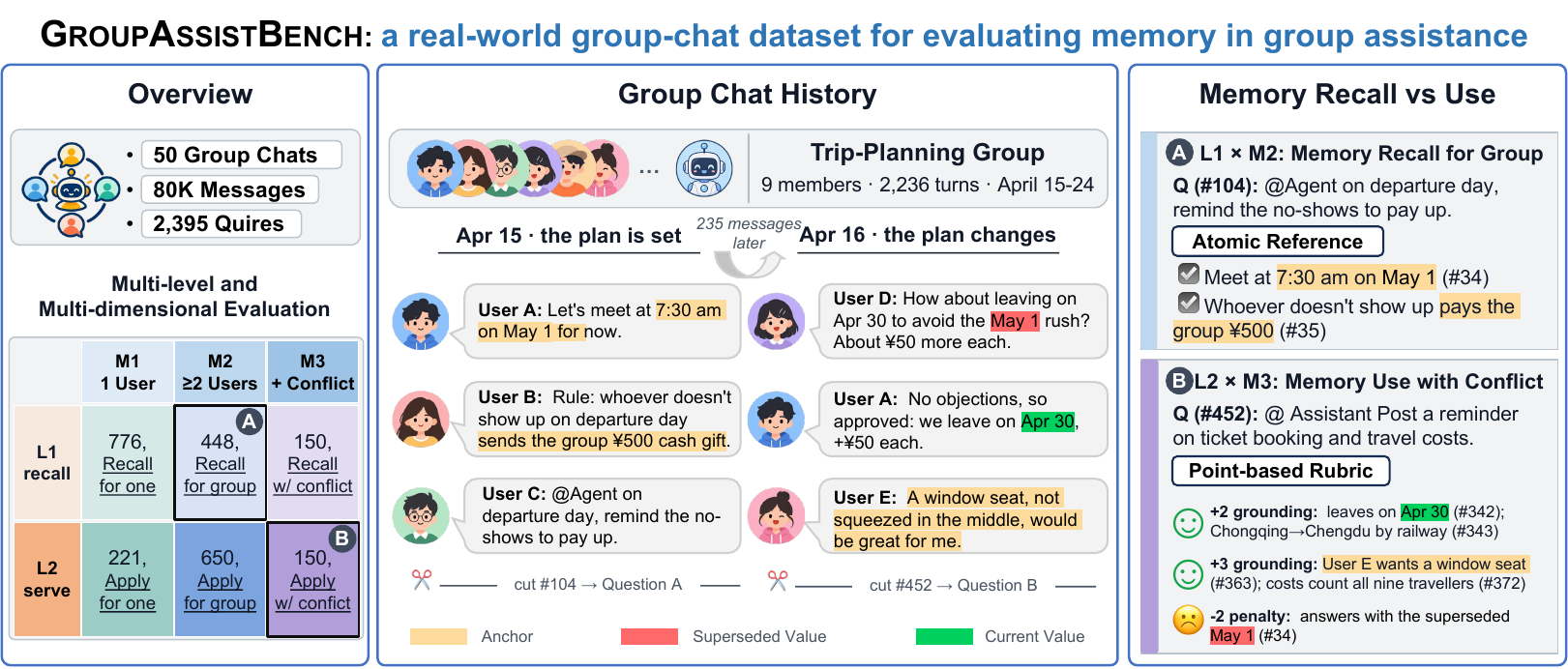}
\caption{Overview of \bench. \textbf{Left:} two task levels, memory recall (L1) and memory use (L2), crossed with three types of memory dependency (M1--M3). \textbf{Middle:} a thread from a trip-planning group. \textbf{Right:} a recall question and a use request derived from the same thread with different history cutoffs: the fact that answers the former is a superseded value for the latter. Messages are translated and abridged.}
\label{fig:overview}
\end{figure}

Existing benchmarks for group memory already examine how information is attributed, integrated, and updated, and show that systems designed for one user exhibit distinct failures in group settings~\citep{DBLP:journals/corr/abs-2605-14498,DBLP:journals/corr/abs-2605-17789}. 
Yet their evaluation inherits a broader limitation of current memory benchmarks: success is measured primarily through recall. Questions ask who said what or which value is current and are scored against fixed reference answers, while tasks closer to assistance use multiple choice~\citep{10.1145/3770855.3817589} or selection from a closed set of actions~\citep{DBLP:journals/corr/abs-2606-18829}. 
This leaves unanswered whether a system that retrieves the right memory can use it to respond appropriately to a service request without a fixed answer. Evidence from settings with one user shows why this distinction matters: on AlpsBench, memory raises persona awareness from 0.41 to 0.72 while reducing emotional appropriateness below the backbone without memory~\citep{xiao2026alpsbench}. 
A separate limitation is that current benchmarks for group memory use synthetic conversations, leaving it unclear whether their tasks capture how agreements, corrections, and assistance failures arise in deployed groups~\citep{DBLP:journals/corr/abs-2605-17789,10.1145/3770855.3817589}.

In this paper, we introduce \bench, a benchmark built on 50 group conversations involving a deployed assistant, containing 80K messages and 2,395 questions. 
\bench organises evaluation along two orthogonal axes, include: 1) The task axis, aims to distinguish memory recall (L1), which asks a question about the history and has a reference answer, from memory use (L2), which poses a service request without a fixed answer and requires a reply that respects the memories it depends on. This axis measures whether a model can move from retrieving remembered information to using it appropriately in an open-ended response, and 2) The dependency axis, aims to distinguish whether the required memories are single (M1), distributed across members (M2), or conflicting (M3). This axis measures how well a model can integrate and reconcile memory evidence as its dependency structure becomes more complex. Figure~\ref{fig:overview} gives an overview of \bench. 
Each request is verified to require the group history, and each required memory is anchored to a verbatim span in the original conversation. 
L1 answers are scored on atomic facts, while L2 replies are scored with rubrics tailored to each instance, whose penalty criteria are derived from real user feedback on the assistant. 
As shown in Table~\ref{tab:comparison}, \bench is the first benchmark that combines conversations among multiple participants with a deployed assistant, memory use through requests without fixed answers, and criteria derived from feedback.

We evaluate 14 configurations on \bench, covering frontier LLMs with full context (e.g., Claude-sonnet-5 and GPT-5.6-terra), retrieval-augmented systems, and external memory systems. 
The results reveal a gap along each axis. Along the task axis, the strongest memory system surpasses Full Context with the same backbone on memory recall (0.621 vs. 0.570) but falls far behind on memory use (0.247 vs. 0.431). Along the dependency axis, the Full Context score for memory use drops from 0.505 on M1 to 0.303 on M3. 
Further analysis highlights two promising directions for group memory assistants: relational evidence organisation and assistance constraint binding (\S\ref{sec:analysis}). Together, these results show that evaluating recall alone overstates the value of memory to a group.
Our contributions are threefold:
\begin{enumerate}
    \item We introduce \bench, the first benchmark built on real group conversations to evaluate memory along two axes: recall versus use, and single versus multiple or conflicting memories.
    \item We develop a traceable framework that verifies history dependence, grounds memories in source messages, and evaluates responses using criteria specific to each request and based on real user feedback.
    \item Experiments across full context, retrieval methods, and memory systems reveal complementary retrieval and application bottlenecks that relational evidence organisation and assistance constraint binding mitigate.
\end{enumerate}

\begin{table}[h]
\centering
\caption{Comparison of \bench with existing memory benchmarks. \emph{Open service} denotes generation for service requests without fixed answers. \emph{Rubric} and \emph{Feedback} indicate criteria tailored to each instance and criteria derived from real user feedback. \emph{\#Eval.}is the number of evaluated questions or checkpoints. \pmark\ denotes partial coverage, including constrained outputs such as multiple choice, template filling, or a fixed action set.}
\label{tab:comparison}
\small
\setlength{\tabcolsep}{3pt}
\resizebox{\linewidth}{!}{%
\begin{tabular}{lccccccr}
\toprule
Benchmark & Group & Source & \shortstack{Update or\\conflict} & \shortstack{Open\\service} & Rubric & Feedback & \#Eval. \\
\midrule
\multicolumn{8}{l}{\emph{Memory for one user and agent tasks}} \\
MemoryArena~\citep{he2026benchmarking}                    & \xmark & partly synthetic & \xmark & \pmark & \xmark & \xmark & 766 \\
AMA-Bench~\citep{zhao2026amabench}                        & \xmark & agent trajectories & \cmark & \xmark & \xmark & \xmark & 2,496 \\
PersonaMem-v2~\citep{DBLP:journals/corr/abs-2512-06688}   & \xmark & synthetic        & \cmark & \pmark & \xmark & \xmark & 5,000 \\
AlpsBench~\citep{xiao2026alpsbench}                       & \xmark & real user logs   & \pmark & \cmark & \xmark & \xmark & 2,500 \\
DynamicMem~\citep{DBLP:journals/corr/abs-2606-22877}      & \xmark & synthetic        & \cmark & \pmark & \xmark & \xmark & 3,634 \\
\midrule
\multicolumn{8}{l}{\emph{Memory for groups}} \\
GroupMemBench~\citep{DBLP:journals/corr/abs-2605-14498}   & \cmark & synthetic & \cmark & \xmark & \xmark & \xmark & 715 \\
EverMemBench~\citep{10.1145/3770855.3817589}              & \cmark & synthetic & \cmark & \pmark & \xmark & \xmark & 2,400 \\
SocialMemBench~\citep{DBLP:journals/corr/abs-2605-17789}  & \cmark & synthetic & \cmark & \xmark & \xmark & \xmark & 1,031 \\
GateMem~\citep{DBLP:journals/corr/abs-2606-18829}         & \cmark & synthetic & \pmark & \pmark & \cmark & \xmark & 2,218 \\
\midrule
\textbf{\bench (ours)} & \cmark & real group chats & \cmark & \cmark & \cmark & \cmark & 2,395 \\
\bottomrule
\end{tabular}%
}
\end{table}

%% file: sections/2_related_work_v4.tex
\section{Related Work}

\paragraph{Memory Evaluation from Recall to Use.}
Long-term memory benchmarks began with factual recall over extended conversation histories~\citep{maharana-etal-2024-evaluating,wu2025longmemeval,hu2026evaluating} and have since turned to how memory supports downstream behavior, such as personalized responses~\citep{DBLP:journals/corr/abs-2512-06688,xiao2026alpsbench,DBLP:journals/corr/abs-2606-22877}, implicit constraints on later replies~\citep{li-etal-2026-locomo}, and multi-session agentic tasks~\citep{he2026benchmarking,zhao2026amabench}. All of them target a single user and score memory use in a closed form, or judge free-form answers holistically~\citep{wu2025longmemeval,li-etal-2026-locomo}, which does not tell which memory was used or misused. Rubric-based evaluation is finer grained~\citep{DBLP:journals/corr/abs-2505-08775,ding2026adarubric}, but its criteria are written by benchmark designers, and real user feedback has served only as a satisfaction label on the whole response~\citep{hong2026uxbenchbenchmarkinguserexperience}. \bench scores open-ended memory use whose required memories come from multiple users, with grounding criteria instantiated from the verified memories of each request and penalty criteria derived from real user feedback on the assistant's failures, so that the criteria name which memory was misused.

\paragraph{Memory in Multi-Party Conversations.}
Recent benchmarks extend memory evaluation to groups, covering speaker-grounded belief tracking and group dynamics~\citep{DBLP:journals/corr/abs-2605-14498}, temporally evolving decisions across interleaved groups~\citep{10.1145/3770855.3817589}, group norms versus individual exceptions~\citep{DBLP:journals/corr/abs-2605-17789}, and memory governance among multiple principals~\citep{DBLP:journals/corr/abs-2606-18829}. They show that memory systems designed for a single user fail characteristically in groups, for example by conflating speakers or overwriting earlier statements. As Table~\ref{tab:comparison} summarizes, their conversations are LLM-synthesized and their tasks are scored against fixed answers, with assistance-like tasks cast as multiple-choice questions~\citep{10.1145/3770855.3817589} or a closed set of actions~\citep{DBLP:journals/corr/abs-2606-18829}; where per-instance criteria exist, they specify required content~\citep{DBLP:journals/corr/abs-2606-18829} or penalize misattribution~\citep{DBLP:journals/corr/abs-2605-17789} by design rather than from observed user reactions. \bench instead builds on conversations with a deployed assistant and evaluates recall and use over the same histories.

%% file: sections/3_methodology_v3.tex
\section{\bench}
\label{sec:bench}

\bench evaluates whether a system can recall a group's history and, more importantly, whether it can use that history to assist the group. This section formulates the tasks (Section~\ref{sec:task}), presents the construction pipeline that turns group conversations into instances (Section~\ref{sec:construction}), and defines the scoring of responses (Section~\ref{sec:evaluation}).

\subsection{Task Formulation}
\label{sec:task}

A group conversation is a chronologically ordered sequence $H=(m_1,\dots,m_T)$, where each message $m_t=(u_t,\tau_t,x_t)$ consists of a speaker $u_t$, who is either a group member or the assistant, a timestamp $\tau_t$, and a text $x_t$. 
An instance of \bench is a tuple $(c, q, A, K)$. The cutoff $c$ specifies the history $H_{<c}$ that a system may access, and $q$ is a query posed at the cutoff. $A=\{a_1,\dots,a_n\}\subset H_{<c}$ is the set of \emph{required memories}, the facts established in the history that a good response to $q$ depends on, each anchored to a message in $H_{<c}$. $K$ is the answer key, a set of reference facts for memory recall or a rubric for memory use. 
Given $H_{<c}$ and $q$, a system produces a response $\hat{y}$, and its goal is to score as high as possible under $K$. $A$ is never shown to the system: it determines the dependency type of the instance, generates $K$, and allows the evidence a system retrieves to be checked against the memories it should have found. 
Every instance is labeled along two orthogonal axes, the \emph{task level} (L1 or L2), which fixes the form of $q$ and $K$, and the \emph{memory dependency} (M1 to M3), which describes the structure of $A$.

\paragraph{Task axis.} 
In \emph{memory recall} (L1), $q$ is a question about the history itself, such as which restaurant the group has settled on, and $K$ is a set of reference facts, one for each memory in $A$. 
In \emph{memory use} (L2), $q$ is an open-ended request for assistance, such as drafting a plan, making a recommendation, or writing a summary for the group, and $K$ is a rubric specific to the instance, since the memories in $A$ constrain an appropriate response without determining a single output.
For the reminder request in Figure~\ref{fig:overview}, for example, the rubric checks that the reply uses the revised departure date and counts all nine travellers, and penalizes reverting to the original date.
Both levels are built over the same group histories, which allows us to compare how well a system recalls the memories with how well it uses them.

\paragraph{Memory axis.}  
\emph{Single memory} (M1) instances depend on a single substantive memory. 
\emph{Multiple memories} (M2) instances depend on multiple memories that must be combined, which in a group are typically contributed by different members. 
\emph{Conflicting memories} (M3) instances have the structure of M2 and additionally contain a \emph{conflict pair} $(a^{\mathrm{old}}, a^{\mathrm{new}})\subset A$: two values of the same matter, stated at different times or by different members, of which the conversation has made $a^{\mathrm{new}}$ the current one. 
The query gives no cue that anything has changed, so a system must notice the revision on its own; the departure date of Figure~\ref{fig:overview}, moved from May 1 to April 30, is such a pair.
These types describe the structure of memory dependency rather than a guaranteed order of difficulty.

\begin{figure}[t]
\centering
\includegraphics[width=\linewidth]{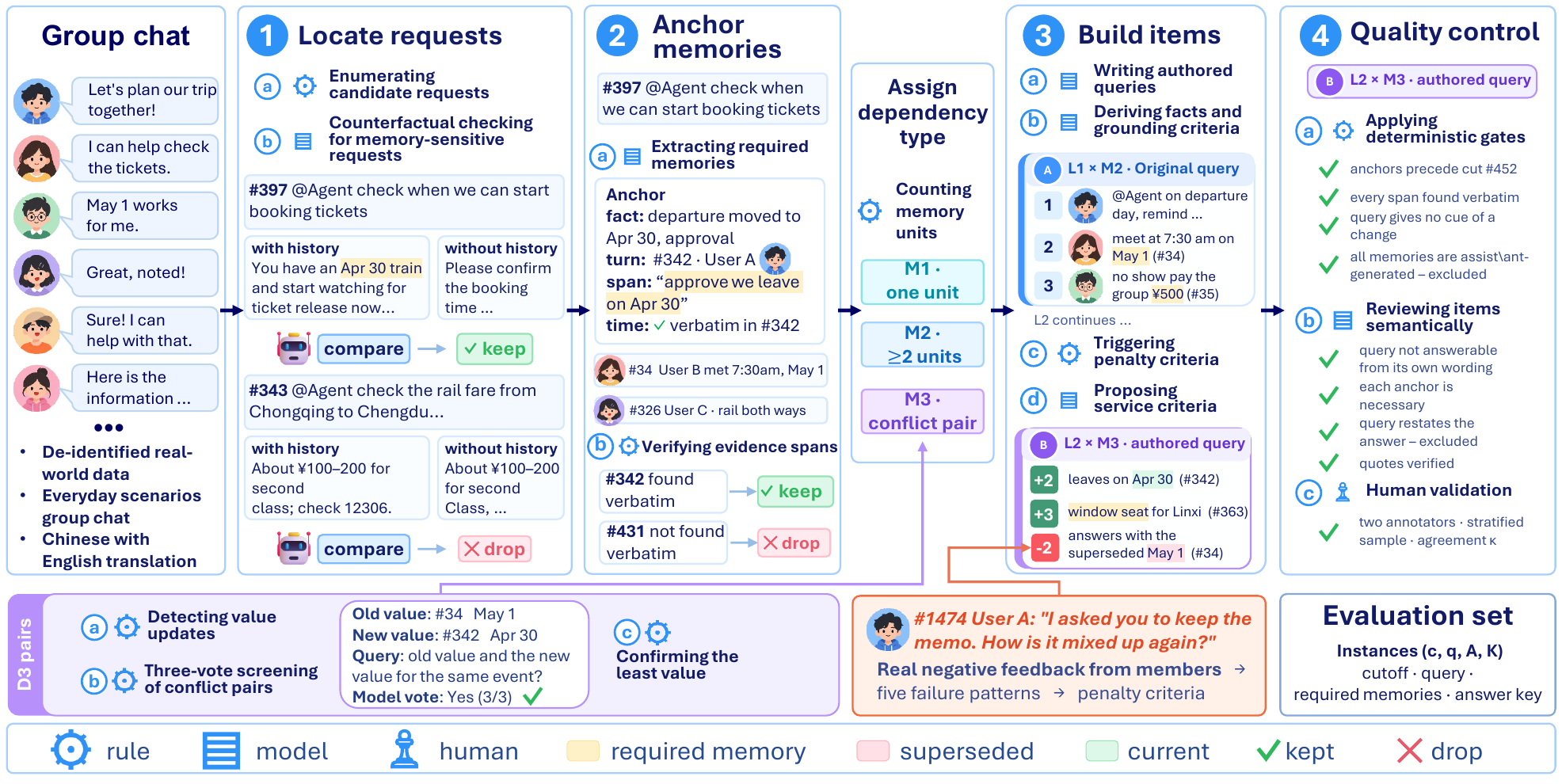}
\caption{Construction pipeline of \bench from group conversations to evaluation instances. Requests are retained only when the group history changes the answer. Required memories are anchored to verbatim spans, and their structure determines the dependency type. Instances use reference facts for L1 or rubrics for L2, with penalties derived from real user feedback, before passing deterministic gates and semantic review. Icons indicate whether each step uses rules, a language model, or humans. Appendix~\ref{app:construction} provides details.}
\label{fig:pipeline}
\end{figure}

\subsection{Benchmark Construction}
\label{sec:construction}

The construction pipeline of \bench is shown in Figure~\ref{fig:pipeline}. 
We \emph{locate} the requests whose answer depends on the group history, \emph{anchor} the memories each request requires to the messages that establish them, \emph{assign} the dependency type from the structure of the anchors, \emph{build} the instance with its answer key, and apply \emph{quality control}. 
Throughout, every field produced by a language model is checked programmatically before it enters the benchmark, and instances that fail are excluded rather than repaired.

\paragraph{Data source and statistics.}
\bench is built from Chinese group conversations in a deployed application, and the assistant replies are its original outputs. Messages are pseudonymized and rewritten to remove residual personal information, with English translations provided. The groups span everyday settings such as trip planning, classes, hobby communities, and project coordination. The evaluation set contains 2,395 instances sampled from 9.5K validated candidates across 50 groups and 80K messages. Although M2 is defined by the number of required memories, 56\% of M2 and 93\% of M3 instances draw anchors from at least two members. Table~\ref{tab:stats} reports the full statistics. Appendix~\ref{app:source} details source selection, and Appendix~\ref{sec:appendix_case} provides more examples.

\paragraph{Locating requests.} 
A request enters the benchmark only if answering it requires the group history. 
We enumerate the messages that address the assistant, discard those that are not requests, and answer each remaining request $q$ posed at message $t_q$ twice with the same model, with and without the preceding history, $\hat{y}^{+}=f(H_{<t_q}, q)$ and $\hat{y}^{-}=f(\varnothing, q)$. 
A comparator keeps $q$ only if $\hat{y}^{+}$ contains group-specific information that $\hat{y}^{-}$ cannot supply and that materially improves the response. 
A retained request becomes the query of a \emph{native} instance, with the cutoff $c=t_q$.

\paragraph{Anchoring memories.} 
For each retained request we identify the minimal set of memories needed to respond. 
Each memory is recorded as an \emph{anchor} $a=(f,t,e)$: a self-contained statement of the fact $f$, the message $t$ that establishes it, and an evidence span $e\sqsubseteq x_t$, a verbatim substring of that message. 
An extractor proposes anchors and drops any whose removal would not change the response. 
The verified anchors of all requests form a per-group inventory, from which authored instances are later built.

\paragraph{Assigning dependency types.} 
The dependency type describes the structure of $A$. Anchors from the same message and near-duplicate facts are merged into memory units; an instance is M1 if it depends on one unit and M2 if on two or more. 
M3 additionally requires a verified conflict pair $(a^{\mathrm{old}},a^{\mathrm{new}})\subset A$. 
Candidate pairs are detected from update expressions and value changes in the history and matched to the anchor of the earlier value. Since two similar values may concern different matters, e.g., a lodging budget and a ticket price, each candidate receives three independent votes on whether the values concern the same matter, and only unanimously accepted pairs are used. 
Finally, we verify that $a^{\mathrm{new}}$ is still current at the cutoff, and move the cutoff before the next change of the same matter if it is not.

\paragraph{Building instances.} 
Native requests cover only the memories that members happened to ask about. 
To cover the remaining anchors, we also build \emph{authored} instances, in which a query is written for a set of anchors $A$ drawn from the inventory. 
Every factual premise in the query must cite a message $t$ and a span $e\sqsubseteq x_t$, verified as for anchors, and the cutoff is placed between $\delta_{\min}$ and $\delta_{\max}$ messages after the last anchor, with no update of the same matter in between. The answer key is then derived from $A$. For L1, $K$ is the set of reference facts $\{f_a : a\in A\}$, one per anchor. 
For L2, $K$ is a rubric assembled from three kinds of criteria,
\begin{equation}
K \;=\; R_{\mathrm{ground}}(A)\;\cup\;R_{\mathrm{pen}}(A)\;\cup\;R_{\mathrm{serv}},
\label{eq:rubric-composition}
\end{equation}
where grounding criteria are instantiated from the anchors, penalty criteria are added by deterministic triggers over the anchors and the memory state, and service criteria are proposed by a language model for the request itself.

\paragraph{Quality control.} 
Every instance passes deterministic gates: each anchor and cited premise must precede the cutoff, each span must be found verbatim in its message, and the query must neither reveal a required memory nor, for M3, hint that a value has changed. The instances selected for evaluation additionally pass a language-model review, which checks whether each anchor is necessary, whether the query can be answered from its own wording, whether an anchor in an assistant message relays a fact established by a member, and, for M3, whether the two values concern the same matter; every negative judgment must quote a span that supports it, which is again verified programmatically (Appendix~\ref{app:qc}). 
To assess the resulting quality, three annotators independently examined 120 instances stratified by level, dependency, and source: 95.8\% are valid by majority vote and 84.2\% unanimously, with pairwise agreement of 93--98\% per question. On the 60 L2 instances, the annotators further rated each rubric for the soundness of its service criteria, the relevance of its penalty criteria, and its overall completeness; 90.0\% of the rubrics pass all three checks by majority and 86.7\% unanimously (Appendix~\ref{app:human}).

\begin{table}[t]
\centering
\small
\caption{Statistics of \bench. \textbf{Left:} number of instances by task level and memory dependency. \textbf{Right:} statistics over the 50 groups.}
\label{tab:stats}
\begin{minipage}[t]{0.48\linewidth}
\centering
\begin{tabular}{l rrr r}
\toprule
 & M1 & M2 & M3 & Total \\
\midrule
L1: memory recall & 776 & 448 & 150 & 1,374 \\
L2: memory use    & 221 & 650 & 150 & 1,021 \\
\midrule
Total             & 997 & 1,098 & 300 & 2,395 \\
\bottomrule
\end{tabular}
\end{minipage}\hfill
\begin{minipage}[t]{0.48\linewidth}
\centering
\resizebox{\linewidth}{!}{%
\begin{tabular}{l rrrr}
\toprule
 & Total & Min. & Mean & Max. \\
\midrule
Instances per group  & 2,395  & 4    & 47.9  & 110 \\
Messages per group   & 80,299 & 68   & 1,606 & 4,442 \\
Context tokens       & 4.92M  & 2.9K & 98.5K & 972.8K \\
Anchors per instance & 6,322  & 1    & 2.64   & 59 \\
\bottomrule
\end{tabular}%
}
\end{minipage}
\end{table}

\subsection{Evaluation}
\label{sec:evaluation}

The two levels are scored with different units of judgment, atomic facts for memory recall and rubric criteria for memory use, so that failures of recalling a memory can be separated from failures of using it.

\paragraph{Memory recall.} For an L1 instance $i$, the answer key $K_i$ is a set of reference facts $G_i$. A judge model decides for each fact $g\in G_i$ whether the response $\hat{y}_i$ recovers it, up to paraphrase; let $m_{ig}\in\{0,1\}$ denote this judgment. Atom Recall is the fraction of reference facts recovered,
\begin{equation}
R_i=\frac{1}{|G_i|}\sum_{g\in G_i} m_{ig}.
\label{eq:recall}
\end{equation}
Let $H_i=\sum_{g\in G_i}m_{ig}$ be the number of recovered facts. The judge also counts additional factual points $E_i$, yielding factual precision $P_i=H_i/(H_i+E_i)$ and $F_{1,i}=2P_iR_i/(P_i+R_i)$.

\paragraph{Memory use.} An L2 response is judged against its rubric $K_i$ (Equation~\ref{eq:rubric-composition}). A grounding criterion is met only if the response explicitly reflects the fact of its anchor. Penalty criteria describe failures of memory use that we summarised from negative feedback that members gave to the assistant: distorting a recalled fact, contradicting a settled plan, violating an explicit agreement or a member's constraint, misreading a member's current state, and relying on a superseded value; a penalty criterion is met when the response commits the failure. A judge evaluates each criterion independently. Positive criteria receive weights from $1$ to $3$, with the largest for required memories, while penalties receive weights from $-1$ to $-3$. For binary judgment $m_{ij}$ and signed weight $w_{ij}$ on criterion $j$, the rubric score is
\begin{equation}
S_i=\frac{\sum_{j} w_{ij}\,m_{ij}}{\sum_{j:\,w_{ij}>0} w_{ij}}.
\label{eq:rubric}
\end{equation}
Criteria and reference facts are shown only to the judge; weights and trigger rules are given in Appendix~\ref{app:rubric}.

%% file: sections/4_experiment.tex
\section{Experiments}
\label{sec:experiments}

\subsection{Experimental Setup}
\label{sec:experimental-setup}

\paragraph{Baselines.}

Following prior work \cite{zhao2026amabench,he2026benchmarking}, we compare three memory-access paradigms: \emph{Full Context}, \emph{Retrieval-Augmented Systems}, and \emph{External Memory Systems}. We characterise their memory organisation using the topology taxonomy of \citet{DBLP:journals/corr/abs-2512-13564}: 1D \emph{flat} memory has no explicit inter-unit relations, 2D \emph{planar} memory organises units within a single structural layer, and 3D \emph{hierarchical} memory connects representations across multiple abstraction levels. Full Context is a 1D reference that directly provides all prior messages and assistant trajectories, and we evaluate four backbones: Claude-sonnet-5, GPT-5.6-terra, Deepseek-v4-flash and Qwen-3.5-9b. Retrieval-Augmented Systems index the same history and include two 1D methods, BM25 and dense embedding retrieval, one 2D method, GraphRAG \cite{DBLP:journals/corr/abs-2404-16130}, and one 3D method, RAPTOR \cite{sarthi2024raptor}. External Memory Systems abstract, organise, or update the interaction history before retrieval. They include two representatives at each level: Mem0 \cite{DBLP:conf/ecai/ChhikaraKASY25} and LightMem \cite{fang2026lightmem} for 1D, A-MEM \cite{xu2025amem} and MemoryOS \cite{kang-etal-2025-memory} for 2D, and xMemory \cite{hu2026beyond} and EverMemOS \cite{hu-etal-2026-evermemos} for 3D. This design covers both retrieval and external memory across the three topology levels without treating structural complexity as an ordering of capability.

\paragraph{Implementation details.}
We use GPT-5.6-terra as the task agent for retrieval and external memory systems and GPT-5.6-sol as the evaluator. BM25 and dense retrieval return the top 20 units. All embedding matching uses \texttt{text-embedding-3-small}. Because the source conversations are in Chinese, we execute and score tasks in Chinese to preserve their meaning and context and avoid translation artefacts. We provide a complete English translation for reference and future research. Each question has a cutoff that limits context and memory construction to earlier turns, preventing answer leakage. Appendices~\ref{app:experimental-details} and~\ref{app:prompt-design} detail the setup and prompts.

\paragraph{Evaluation metrics.}
Using GPT-5.6-sol as the judge, we report mean Atom Recall and factual F1 for L1, mean rubric score $\bar{S}$ for L2, and mean context tokens. We break results down by M1--M3 and rank methods by overall Atom Recall and $\bar{S}$, following Section~\ref{sec:evaluation}. To validate the judge, three annotators independently marked every rubric criterion for 180 real responses to 60 sampled L2 instances. The judge agrees with the human majority on 89.3\% of the 1,152 criterion judgments, the same level as the annotators agree with one another (89.8\%). When errors occur, the judge tends to give too little credit to grounding and service criteria and apply penalties too often, making scores conservative. The judge also ranks the three systems identically to every annotator. Appendix~\ref{app:human-judge} provides details.

\subsection{Main Experiment}

\begin{table*}[t]
\centering
\caption{Main results on \bench{} for memory recall (L1) and memory use (L2). All RAG and external memory systems use GPT-5.6-terra as the task agent. We report Atom Recall (met) and F1 for L1. For L2, $\bar{S}$ denotes the mean of the rubric scores $S_i$. Ranks are based on mean met over all L1 questions and mean $S_i$ over all L2 questions. Bold numerical entries indicate maximum scores across all methods and rank 1. Underlined numerical entries indicate maximum scores and the highest rank within each group.}
\label{tab:main-results-pipeline-0916-300}
\begingroup
\small
\setlength{\tabcolsep}{3pt}
\renewcommand{\arraystretch}{1.18}
\resizebox{\textwidth}{!}{%
\begin{tabular}{@{}lc*{14}{c}@{}}
\toprule
\multirow{2}{*}{\textbf{Method}}
& \multirow{2}{*}{\textbf{\shortstack{Memory\\type}}}
& \multicolumn{2}{c}{\textbf{L1-M1}}
& \multicolumn{2}{c}{\textbf{L1-M2}}
& \multicolumn{2}{c}{\textbf{L1-M3}}
& \multirow{2}{*}{\textbf{\boldmath\shortstack{L1 avg.\\met $\uparrow$}}}
& \multirow{2}{*}{\textbf{\boldmath\shortstack{L1\\rank $\downarrow$}}}
& \textbf{L2-M1} & \textbf{L2-M2} & \textbf{L2-M3}
& \multirow{2}{*}{\textbf{\boldmath\shortstack{L2 avg.\\$\bar{S}\,\uparrow$}}}
& \multirow{2}{*}{\textbf{\boldmath\shortstack{L2\\rank $\downarrow$}}}
& \multirow{2}{*}{\textbf{\shortstack{Ctx.\\tokens}}} \\
\cmidrule(lr){3-4}\cmidrule(lr){5-6}\cmidrule(lr){7-8}
\cmidrule(lr){11-11}\cmidrule(lr){12-12}\cmidrule(lr){13-13}
& & \textbf{\boldmath met $\uparrow$} & \textbf{\boldmath F1 $\uparrow$}
& \textbf{\boldmath met $\uparrow$} & \textbf{\boldmath F1 $\uparrow$}
& \textbf{\boldmath met $\uparrow$} & \textbf{\boldmath F1 $\uparrow$}
& & & \textbf{\boldmath $\bar{S}\,\uparrow$} & \textbf{\boldmath $\bar{S}\,\uparrow$} & \textbf{\boldmath $\bar{S}\,\uparrow$} & & & \\
\midrule
\rowcolor{fullcontextbg}
\multicolumn{16}{@{}l}{\textit{Full Context}} \\
\addlinespace[2pt]
\textbf{Claude Sonnet 5} & 1D & 0.739 & 0.570 & 0.504 & 0.460 & 0.429 & 0.352 & 0.628 & 2 & 0.473 & 0.426 & 0.371 & 0.428 & 3 & 52,435 \\
\textbf{DeepSeek-v4-flash} & 1D & \underline{\textbf{0.766}} & 0.526 & \underline{\textbf{0.572}} & \underline{\textbf{0.472}} & \underline{0.455} & 0.353 & \underline{\textbf{0.669}} & \underline{\textbf{1}} & \underline{\textbf{0.510}} & \underline{\textbf{0.478}} & \underline{\textbf{0.374}} & \underline{\textbf{0.469}} & \underline{\textbf{1}} & 52,435 \\
\textbf{Qwen-3.5-9b} & 1D & 0.622 & 0.467 & 0.344 & 0.341 & 0.285 & 0.278 & 0.494 & 11 & 0.446 & 0.343 & 0.226 & 0.348 & 6 & 52,435 \\
\textbf{GPT-5.6-terra} & 1D & 0.685 & \underline{\textbf{0.623}} & 0.436 & 0.453 & 0.377 & \underline{\textbf{0.356}} & 0.570 & 4 & 0.505 & 0.435 & 0.303 & 0.431 & 2 & 52,435 \\
\midrule
\rowcolor{retrievalbg}
\multicolumn{16}{@{}l}{\textit{Retrieval-Augmented Systems}} \\
\addlinespace[2pt]
\textbf{BM25} & 1D & \underline{0.676} & \underline{0.612} & \underline{0.378} & \underline{0.404} & 0.313 & 0.306 & \underline{0.539} & \underline{6} & \underline{0.446} & \underline{0.376} & 0.230 & \underline{0.370} & \underline{4} & 10,769 \\
\textbf{Dense retrieval} & 1D & 0.651 & 0.595 & 0.374 & 0.397 & \underline{0.358} & \underline{0.351} & 0.529 & 7 & 0.428 & 0.368 & \underline{0.242} & 0.362 & 5 & 9,539 \\
\textbf{GraphRAG} & 2D & 0.618 & 0.554 & 0.356 & 0.384 & 0.356 & 0.343 & 0.504 & 10 & 0.399 & 0.351 & 0.236 & 0.344 & 7 & 18,468 \\
\textbf{RAPTOR} & 3D & 0.647 & 0.581 & 0.350 & 0.376 & 0.345 & 0.329 & 0.517 & 8 & 0.422 & 0.327 & 0.237 & 0.334 & 8 & 14,151 \\
\midrule
\rowcolor{externalmemorybg}
\multicolumn{16}{@{}l}{\textit{External Memory Systems}} \\
\addlinespace[2pt]
\textbf{Mem0} & 1D & 0.464 & 0.408 & 0.252 & 0.264 & 0.234 & 0.223 & 0.370 & 13 & 0.305 & 0.241 & 0.172 & 0.245 & 11 & 1,190 \\
\textbf{LightMem} & 1D & 0.290 & 0.261 & 0.133 & 0.155 & 0.178 & 0.189 & 0.227 & 14 & 0.192 & 0.131 & 0.094 & 0.139 & 14 & 742 \\
\textbf{A-MEM} & 2D & 0.537 & 0.406 & 0.342 & 0.300 & 0.355 & 0.261 & 0.454 & 12 & 0.321 & 0.218 & 0.185 & 0.235 & 12 & 5,387 \\
\textbf{MemoryOS} & 2D & 0.623 & 0.451 & 0.387 & 0.327 & 0.327 & 0.234 & 0.513 & 9 & \underline{0.380} & 0.255 & 0.168 & \underline{0.269} & \underline{9} & 5,203 \\
\textbf{xMemory} & 3D & 0.620 & 0.511 & 0.455 & 0.374 & 0.435 & 0.262 & 0.546 & 5 & 0.133 & 0.223 & 0.220 & 0.203 & 13 & 5,841 \\
\textbf{EverMemOS} & 3D & \underline{0.679} & \underline{0.555} & \underline{0.540} & \underline{0.430} & \underline{\textbf{0.561}} & \underline{0.302} & \underline{0.621} & \underline{3} & 0.137 & \underline{0.284} & \underline{0.253} & 0.247 & 10 & 21,123 \\
\bottomrule
\end{tabular}%
}
\endgroup

\end{table*}

\paragraph{Overall results.}
Table~\ref{tab:main-results-pipeline-0916-300} shows that Full Context methods remain competitive on both L1 and L2 tasks. DeepSeek-v4-flash obtains L1 met of 0.669 and L2 rubric score $\bar{S}$ of 0.469, while GPT-5.6-terra reaches 0.570 and 0.431, respectively. Within RAG systems, BM25 and dense retrieval obtain higher overall scores than GraphRAG and RAPTOR on both tasks, indicating that additional structural complexity does not consistently improve performance in this setting. Among external memory systems, hierarchical methods show promising L1 recall: EverMemOS and xMemory attain the met of 0.621 and 0.546, placing third and fifth among all evaluated baselines. Flat retrieval is more effective within RAG, whereas hierarchical systems show stronger L1 recall within external memory. This pattern is consistent with concerns that summary hierarchies can obscure factual distinctions \cite{hu2026beyond} and highlights hierarchical memory organisation as an emerging direction for supporting reasoning across abstraction levels \cite{DBLP:journals/corr/abs-2512-13564}.

\paragraph{Performance across memory dependencies.}
Performance generally degrades as queries move beyond dependence on a single memory. L1 met decreases from M1 to both M2 and M3 for all 14 evaluated baselines, and L2 shows a similar pattern except for xMemory and EverMemOS. For example, with GPT-5.6-terra under Full Context, L1 met drops from 0.685 on M1 to 0.436 and 0.377 on M2 and M3, while the L2 score decreases from 0.505 to 0.435 and 0.303. These drops highlight the difficulty of integrating complementary memories across speakers or turns in M2 and selecting the current or applicable memories amid updates, conflicts, or contextual constraints in M3. The two 3D external memory systems, EverMemOS and xMemory, show the strongest L1 robustness under this shift, with smaller M1 to M3 declines than every Full Context and RAG configuration. EverMemOS declines from 0.679 to 0.561, while Full Context with GPT-5.6-terra declines from 0.685 to 0.377. xMemory declines from 0.620 to 0.435. EverMemOS also improves on L2 from 0.137 on M1 to 0.284 on M2 and 0.253 on M3. Together, these results highlight the potential of hierarchical external memory organisation to preserve and aggregate evidence across multiple memory anchors.

\begin{wrapfigure}{r}{0.48\linewidth}
    \centering
    \IfFileExists{experiment/figures/cost_break_even.pdf}{%
      \includegraphics[width=\linewidth]{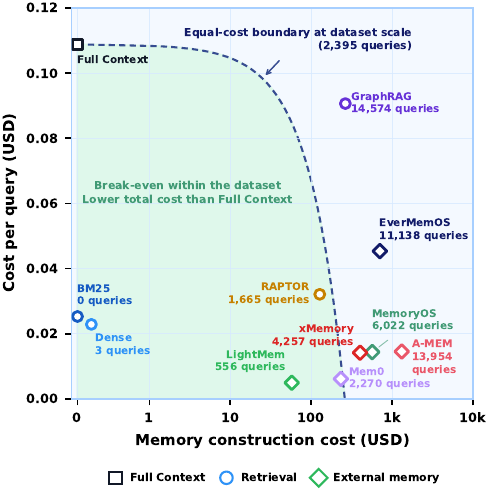}%
    }{%
      \includegraphics[width=\linewidth]{figures/cost_break_even.pdf}%
    }
    \caption{\textbf{Deployment cost efficiency.} The shaded region marks systems that amortize construction within all 2,395 queries relative to Full Context. Labels give the required query count. Judge costs are excluded.}
    \label{fig:cost-break-even}
    \vspace{-2em}
\end{wrapfigure}

\paragraph{Memory recall does not guarantee effective assistance.}
The relative strengths of retrieval and external memory systems reverse from L1 to L2. On L1, EverMemOS surpasses Full Context with GPT-5.6-terra in mean met (0.621 vs. 0.570), while EverMemOS (0.621) and xMemory (0.546) outperform all RAG systems (0.504--0.539). On L2, however, every RAG system (0.334--0.370) outperforms every external memory system (0.139--0.269). This advantage is only relative because all RAG systems remain below Full Context with GPT-5.6-terra (0.431), while BM25 and dense retrieval outperform GraphRAG and RAPTOR. Thus, strong recall by external memory systems does not translate into effective assistance, while RAG provides stronger but still limited support.

\paragraph{End-to-end cost efficiency.}
To examine whether reduced context usage during queries offsets memory construction, we compute total deployment cost over the full dataset as construction cost paid once plus retrieval and generation costs across all queries. Figure~\ref{fig:cost-break-even} separates these components. At the observed workload of all queries, BM25, dense retrieval, LightMem, RAPTOR, and Mem0 cost less than Full Context with the same GPT-5.6-terra task agent. For example, dense retrieval costs \$55.0 in total, compared with \$260.4 for Full Context, while RAPTOR reaches \$204.5 after amortising its \$127.6 construction cost. Lower cost, however, does not imply stronger performance. LightMem costs only \$69.6 but obtains the lowest L1 and L2 averages, whereas EverMemOS achieves the highest L1 average among external memory systems but costs \$814.9 and requires 11,138 queries to break even. Overall, an efficient memory system must balance construction overhead and savings with effective memory recall and assistance. Detailed statistics for each stage are reported in Appendix~\ref{app:efficiency-analysis}.

%% file: sections/5_analysis_v2.tex
\section{Analysis}
\label{sec:analysis}

\begin{figure}[h]
\centering
\includegraphics[width=0.8\linewidth]{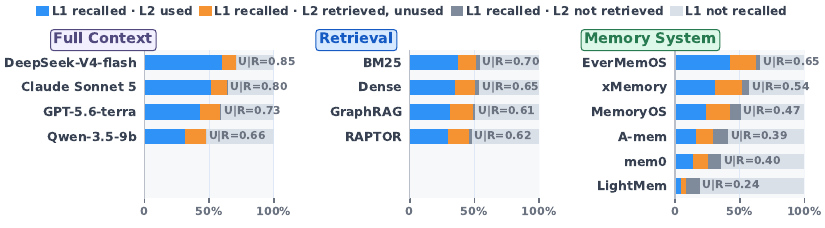}
\vspace{-0.9em}
\caption{\textbf{Fate at L2 of memories recalled at L1.} Each bar covers 497 paired memories. Light gray marks those not recalled at L1. The other segments show whether recalled memories were used, retrieved but unused, or not retrieved at L2. $U\!\mid\!R$ is the blue share among recalled memories, $P(\text{used at L2}\mid\text{recalled at L1})$, not the bar length.}
\label{fig:recall-use}
\vspace{-1.0em}
\end{figure}

\paragraph{Recalled but not used.}
The gap between L1 and L2 does not show whether systems use memories they can recall, since the two levels contain different instances. We therefore identify 497 memories queried directly in L1 and required by an L2 request in the same group, and test whether each system uses in L2 what it recalls in L1 (Figure~\ref{fig:recall-use}). Across all six memory systems, recalled but unused memories account for 21--27\% of all pairs, versus 11--16\% for models given full context. Even EverMemOS matches their recall (66\% vs.\ 47--70\%) but uses fewer recalled memories (65\% vs.\ 73--85\%). This fraction correlates more strongly with the L2 ranking in Table~\ref{tab:main-results-pipeline-0916-300} ($\rho=0.89$) than recall itself does ($\rho=0.56$). Why are memories recalled at L1 left unused at L2? We split unused memories into those absent from the model input and those present but unapplied. (i)~For memory systems, 36\% of unused memories never entered the L2 input. This retrieval failure was cue dependent. L1 questions usually identify the target memory, whereas L2 requests often do not. Four of the six systems retrieved 84--100\% of the same memories when L1 named them directly, and input coverage across all pairs rose from 62\% to 83\% as lexical overlap increased. (ii)~The other 64\% were in the input and still not used in the reply. This failure depends on the cue as well, with memories already in the input used more often when requests shared their wording or were framed around the same matter. Without such cues, memory systems can thus fail twice, in retrieving the relevant memory and in applying it when present. These failures motivate retrieving evidence independently of request wording and binding available memories to the response. Analyses by generation model and request type appear in Appendix~\ref{app:intervention-analysis}.

\begin{finding}
\findlabel{Finding 1}Explicit relational evidence organisation recovers memories that the request does not name.
\end{finding}

\begin{wraptable}{r}{0.52\textwidth}
\centering
\vspace{-2.2em}
\small
\setlength{\tabcolsep}{2.5pt}
\renewcommand{\arraystretch}{1.08}
\caption{\textbf{Relational Evidence Refinement} by memory dependency; Avg weighted by instance counts, green = gain over baseline (pp).}
\label{tab:m-axis-analysis}
\resizebox{\linewidth}{!}{%
\begin{tabular}{@{}ll cc!{\vrule width 0.3pt}cc!{\vrule width 0.3pt}cc@{}}
\toprule
\multicolumn{2}{c}{Metric} & \multicolumn{2}{c}{EverMemOS} & \multicolumn{2}{c}{xMemory} & \multicolumn{2}{c}{LightMem} \\
\cmidrule(lr){3-4}\cmidrule(lr){5-6}\cmidrule(lr){7-8}
& & Baseline & +RER & Baseline & +RER & Baseline & +RER \\
\midrule
\multirow{3}{*}{\shortstack{L1\\met}} & M1 & 0.679 & 0.693\gain{+1.45} & 0.620 & 0.641\gain{+2.11} & 0.290 & 0.380\gain{+8.98} \\
\cmidrule(lr){2-8}
& M2 & 0.540 & 0.544\gain{+0.39} & 0.455 & 0.494\gain{+3.86} & 0.133 & 0.216\gain{+8.13} \\
\cmidrule(lr){2-8}
& M3 & 0.561 & 0.566\gain{+0.50} & 0.435 & 0.456\gain{+2.14} & 0.178 & 0.235\gain{+5.71} \\
\midrule
\multirow{4}{*}{\shortstack{L2\\$\bar{S}$}} & M1 & 0.137 & 0.160\gain{+2.28} & 0.133 & 0.157\gain{+2.32} & 0.192 & 0.215\gain{+2.32} \\
\cmidrule(lr){2-8}
& M2 & 0.284 & 0.294\gain{+1.08} & 0.223 & 0.245\gain{+2.22} & 0.131 & 0.207\gain{+7.52} \\
\cmidrule(lr){2-8}
& M3 & 0.253 & 0.294\gain{+4.04} & 0.220 & 0.238\gain{+1.83} & 0.094 & 0.161\gain{+6.71} \\
\cmidrule(lr){2-8}
& Avg & 0.247 & 0.265\gain{+1.77} & 0.203 & 0.225\gain{+2.19} & 0.139 & 0.202\gain{+6.28} \\
\bottomrule
\end{tabular}%
}
\vspace{-2.0em}
\end{wraptable}

Better generation cannot recover evidence omitted by retrieval. We introduce \textbf{Relational Evidence Refinement (RER)}, which makes relations among retrieved memories explicit and follows them to supplement missing supporting evidence without relying on the wording of the request. We evaluate RER on xMemory and EverMemOS because of their leading L1 performance, and on flat LightMem to test generality. Missing evidence accounts for 74\% of unused memories in LightMem, compared with 22\% in xMemory and 14\% in EverMemOS, as shown by the dark grey segments in Figure~\ref{fig:recall-use}. Table~\ref{tab:m-axis-analysis} confirms this pattern. RER gives flat LightMem the largest average gains (+8.4 L1 met and +6.3 L2 $\bar{S}$), because it omits most evidence, and smaller gains to hierarchical xMemory and EverMemOS, whose evidence is more complete. On L1 M2/M3, anchor coverage exceeds met for xMemory (0.64--0.68 vs.\ 0.44--0.46) and EverMemOS (0.80--0.85 vs.\ 0.54--0.56). RER narrows this gap because integrating evidence across members and turns requires the connecting memories to be present.

\begin{finding}
\findlabel{Finding 2}Effective assistance requires binding memories that are already present to the response.
\end{finding}

\begin{wraptable}{r}{0.52\textwidth}
\centering
\vspace{-2.2em}
\small
\setlength{\tabcolsep}{2.5pt}
\renewcommand{\arraystretch}{1.08}
\caption{\textbf{Assistance Constraint Binding} on L2 by memory dependency.}
\label{tab:l-axis-analysis}
\resizebox{\linewidth}{!}{%
\begin{tabular}{@{}l cc!{\vrule width 0.3pt}cc!{\vrule width 0.3pt}cc@{}}
\toprule
\multirow{2}{*}{\shortstack{L2\\metric}} & \multicolumn{2}{c}{EverMemOS} & \multicolumn{2}{c}{xMemory} & \multicolumn{2}{c}{LightMem} \\
\cmidrule(lr){2-3}\cmidrule(lr){4-5}\cmidrule(lr){6-7}
& Baseline & +ACB & Baseline & +ACB & Baseline & +ACB \\
\midrule
M1 $\bar{S}$ & 0.137 & 0.203\gain{+6.63} & 0.133 & 0.201\gain{+6.77} & 0.192 & 0.212\gain{+1.96} \\
\midrule
M2 $\bar{S}$ & 0.284 & 0.306\gain{+2.17} & 0.223 & 0.257\gain{+3.39} & 0.131 & 0.159\gain{+2.80} \\
\midrule
M3 $\bar{S}$ & 0.253 & 0.276\gain{+2.30} & 0.220 & 0.243\gain{+2.17} & 0.094 & 0.106\gain{+1.20} \\
\midrule
Avg. $\bar{S}$ & 0.247 & 0.279\gain{+3.16} & 0.203 & 0.242\gain{+3.94} & 0.139 & 0.163\gain{+2.38} \\
\bottomrule
\end{tabular}%
}
\vspace{-1.0em}
\end{wraptable}

The larger share of the loss comes from memories that reach the L2 input but remain unused. They account for 78\% of unused memories for xMemory and 86\% for EverMemOS. This application failure helps explain why EverMemOS falls from third on L1 to tenth on L2 and xMemory from fifth to thirteenth. We introduce \textbf{Assistance Constraint Binding (ACB)}, which determines each remembered fact's subject, scope, status, and applicability, then binds the validated conditions to generation. Table~\ref{tab:l-axis-analysis} again matches the decomposition: ACB improves xMemory and EverMemOS most on M1 (+6.8 and +6.6 $\bar{S}$), where the relevant facts are already available and the experiment isolates whether they are interpreted and enforced, while LightMem gains less because its primary bottleneck is missing evidence, which ACB cannot recover. When combined, the two modules perform best for every system on both levels, but their gain is smaller than the sum of their separate gains because each unused memory falls under only one bottleneck. Appendix~\ref{app:intervention-analysis} provides the component analysis with confidence intervals and details of both methods.

%% file: sections/6_conclusion.tex
\section{Conclusion}

We presented \bench, a benchmark built from real-world group conversations with a deployed assistant. It evaluates memory recall and use over the same histories when the required memories are single, multiple, or conflicting. Required memories are anchored to source messages, and L2 rubrics translate real user feedback into criteria tailored to each request. Across LLMs given full context, retrieval methods, and memory systems, recall did not reliably translate into use. The strongest memory system surpassed the same backbone given full context on recall but lagged far behind on use, which declined further with multiple and conflicting memories. Relational Evidence Refinement and Assistance Constraint Binding improved all three tested memory systems by addressing complementary failures in retrieving relevant memories and applying those already available. We hope \bench shifts memory evaluation from what systems recall to what they enable a group to do.

%% file: sections/appendix_v2.tex
\clearpage
\appendix
\providecommand{\bench}{[Dataset Name]}

\begingroup
\hypersetup{linkcolor=black}
\section*{Appendix}
\noindent
\textbf{\hyperref[app:construction]{\ref*{app:construction}\quad Construction Details}} \dotfill \pageref{app:construction}\\[2pt]
\hspace*{1.5em}\hyperref[app:source]{\ref*{app:source}\quad Data Source and Processing} \dotfill \pageref{app:source}\\[2pt]
\hspace*{1.5em}\hyperref[app:locate]{\ref*{app:locate}\quad Locating Requests} \dotfill \pageref{app:locate}\\[2pt]
\hspace*{1.5em}\hyperref[app:anchor]{\ref*{app:anchor}\quad Anchoring Memories} \dotfill \pageref{app:anchor}\\[2pt]
\hspace*{1.5em}\hyperref[app:dependency]{\ref*{app:dependency}\quad Assigning Dependency Types} \dotfill \pageref{app:dependency}\\[2pt]
\hspace*{1.5em}\hyperref[app:build]{\ref*{app:build}\quad Building Instances} \dotfill \pageref{app:build}\\[2pt]
\hspace*{1.5em}\hyperref[app:qc]{\ref*{app:qc}\quad Quality Control} \dotfill \pageref{app:qc}\\[4pt]
\textbf{\hyperref[app:rubric]{\ref*{app:rubric}\quad Rubric Details}} \dotfill \pageref{app:rubric}\\[4pt]
\textbf{\hyperref[app:human]{\ref*{app:human}\quad Human Validation}} \dotfill \pageref{app:human}\\[4pt]
\textbf{\hyperref[app:evaluation-scope]{\ref*{app:evaluation-scope}\quad Dataset Statistics}} \dotfill \pageref{app:evaluation-scope}\\[4pt]
\textbf{\hyperref[app:experimental-details]{\ref*{app:experimental-details}\quad Implementation Setup}} \dotfill \pageref{app:experimental-details}\\[2pt]
\hspace*{1.5em}\hyperref[app:history-cutoffs]{\ref*{app:history-cutoffs}\quad History Cutoffs} \dotfill \pageref{app:history-cutoffs}\\[2pt]
\hspace*{1.5em}\hyperref[app:method-configurations]{\ref*{app:method-configurations}\quad Retrieval and Memory Configuration} \dotfill \pageref{app:method-configurations}\\[2pt]
\hspace*{1.5em}\hyperref[app:generation]{\ref*{app:generation}\quad Generation and Output Validation} \dotfill \pageref{app:generation}\\[4pt]
\textbf{\hyperref[app:experiments]{\ref*{app:experiments}\quad Experiments}} \dotfill \pageref{app:experiments}\\[2pt]
\hspace*{1.5em}\hyperref[app:efficiency-analysis]{\ref*{app:efficiency-analysis}\quad Cost and Token Efficiency} \dotfill \pageref{app:efficiency-analysis}\\[2pt]
\hspace*{1.5em}\hyperref[app:intervention-analysis]{\ref*{app:intervention-analysis}\quad Intervention Analysis} \dotfill \pageref{app:intervention-analysis}\\[2pt]
\textbf{\hyperref[sec:appendix_case]{\ref*{sec:appendix_case}\quad Case Studies}} \dotfill \pageref{sec:appendix_case}\\[4pt]
\textbf{\hyperref[app:prompt-design]{\ref*{app:prompt-design}\quad Prompting and Evaluation Protocol}} \dotfill \pageref{app:prompt-design}\\[2pt]
\hspace*{1.5em}\hyperref[app:prompt-scope]{\ref*{app:prompt-scope}\quad Task and Evaluation Protocol} \dotfill \pageref{app:prompt-scope}\\[2pt]
\hspace*{1.5em}\hyperref[app:answer-prompts]{\ref*{app:answer-prompts}\quad Answer Generation Prompts} \dotfill \pageref{app:answer-prompts}\\[2pt]
\hspace*{1.5em}\hyperref[app:judge-prompts]{\ref*{app:judge-prompts}\quad Evaluation Prompts} \dotfill \pageref{app:judge-prompts}\\[4pt]
\textbf{\hyperref[app:cost-accounting]{\ref*{app:cost-accounting}\quad Cost Accounting}} \dotfill \pageref{app:cost-accounting}\\[4pt]
\par
\endgroup
\clearpage

\section{Construction Details}
\label{app:construction}

This appendix expands the construction pipeline in Section~\ref{sec:construction} and follows the same order as Figure~\ref{fig:pipeline}. Unless otherwise stated, the counts refer to the full validated pool over all groups, from which the evaluation set in Table~\ref{tab:stats} is sampled. All model calls in the pipeline use GPT-5.5, and their cost is reported in Appendix~\ref{app:cost-accounting}. 

\subsection{Data Source and Processing}
\label{app:source}

\paragraph{Selection.} From the group conversations available to us we keep those with at least two members, sustained use of the assistant, and a complete message timeline, which yields 50 groups. Every group is used in full; no segment is removed for evaluation purposes.

\paragraph{Processing.} All conversations are de-identified before they reach the authors. Member identifiers are replaced with per-group pseudonyms, message texts are rewritten to remove residual personal information while preserving the facts, values, and speaker of each message, and a small fraction of messages (2.5\%) are replaced by placeholders of a safety filter. Non-text messages such as images and files are kept as placeholders that name the attachment. Each message keeps its speaker, timestamp, reply relation, and whether it is sent by a member or the assistant. The language of evaluation is described in Appendix~\ref{app:prompt-scope}.

\subsection{Locating Requests}
\label{app:locate}

A \emph{service point} is a message that mentions the assistant or immediately precedes an assistant reply. A lightweight classifier discards service points that are not requests, such as small talk or comments on a previous reply. Each remaining request is answered twice by the same model, once with the preceding history and once without it, and a comparator labels the contribution of the history as \emph{strong}, \emph{weak}, or \emph{none}. Requests labelled \emph{none} are discarded. Information that only appears in earlier assistant replies is excluded when judging the contribution, since it is generated content rather than memory of the group. Of 13,782 service points, 10,191 are requests and 7,926 are retained (6,772 strong and 1,154 weak).

\subsection{Anchoring Memories}
\label{app:anchor}

Anchors are proposed in one call with two passes. The first pass extracts candidate facts that the response depends on, and the second pass removes candidates whose absence would not change the response. Each anchor must state its fact without pointer expressions such as ``the time mentioned above'', and anchors that contain such expressions are rejected by a word list. The evidence span of each anchor is matched verbatim against the cited message, and anchors that fail are rejected (154 of 22,843 proposals). When a fact appears repeatedly, a rescan of the history anchors it to the earliest message that establishes it, and earlier lexical matches are flagged since a lexical match does not necessarily establish the same fact. Anchors whose source message is an assistant message are marked, and the reviewer in Appendix~\ref{app:qc} checks whether they relay a fact established by a member. Verified anchors are tracked in a per-group inventory that records which instances consume each anchor, which is used to select uncovered anchors for authored instances.

\subsection{Assigning Dependency Types}
\label{app:dependency}

\paragraph{M1 and M2.} Anchors from the same message and near-duplicate facts are merged into one \emph{memory unit}, and anchors that only carry the profile of a member do not count. An instance with one unit is M1, and an instance with two or more units is M2.

\paragraph{M3.} Candidate conflict pairs are detected from update expressions and from changes of the value in the same slot, and each new value is paired with the anchor that established the earlier value. Disagreements between members are included only if the conversation later resolves them. Each candidate pair receives three independent votes on the relation between the two values, including \emph{true update}, \emph{different slot}, \emph{range extension}, \emph{rephrase}, and \emph{goal versus status}. Pairs with three \emph{true update} votes are accepted, and the other pairs are not used unless a human annotator confirms them. Of 595 candidate pairs, 338 are accepted. For each accepted pair, we check that the new value is still current at the cutoff. If the matter changes again later, the cutoff is moved before the next change, or the pair is dropped. A M3 instance further requires at least two anchors on the same matter from members other than the one who makes the update, so that it keeps the structure of M2. The superseded value is excluded from the grounding criteria and enters the penalty criteria instead (Appendix~\ref{app:rubric}).

\subsection{Building Instances}
\label{app:build}

Native instances keep the original request of a member and place the cutoff at the request message. For authored instances, the query type is assigned by the pipeline in rotation from a list of nine types: single-fact recall, multi-entity attribute mapping, attribution, state tracking, constraint compliance, update resolution, aggregation, personalized assistance, and arbitration among members. Every factual premise in an authored query must cite a source message and a matching span, which is verified in the same way as an anchor. The cutoff of an authored instance is placed between $\delta_{\min}=$ 20 and $\delta_{\max}=$ 200 messages after its last anchor (Section~\ref{sec:construction}), so that the instances cover a range of distances between a memory and its use, and no update of the same matter may fall within this range. How the cutoff restricts the input of each system at evaluation time is described in Appendix~\ref{app:history-cutoffs}. For L1 instances, the reference answer is split into atomic facts, one for each anchor, and for M3 instances the superseded value is recorded as a distractor. Rubrics of L2 instances are assembled as described in Section~\ref{sec:evaluation} and Appendix~\ref{app:rubric}.

\subsection{Quality Control}
\label{app:qc}

\paragraph{Deterministic gates.} An instance is rejected if (i) any anchor or cited premise does not precede the cutoff, (ii) any evidence span is not found verbatim in the cited message, (iii) the query reveals a required memory, names an evaluation dimension, or, for M3 instances, contains words that hint at a change, (iv) the cutoff leaves almost no history, or (v) the rubric violates the constraints in Appendix~\ref{app:rubric}.

\paragraph{Language-model review.} A reviewer model then checks (i) whether each anchor is necessary, by testing whether a smaller set of anchors is sufficient, (ii) whether the query can be answered from its own wording, (iii) whether an anchor in an assistant message relays a fact established by a member or content generated by the assistant, and (iv) whether the two values of a M3 instance belong to the same matter. For every negative judgment the reviewer must quote the span that supports it, and quotes that are not found in the conversation invalidate the judgment. Instances that fail either layer are excluded from the pool but kept in the audit log with the reason. In total, 9,542 instances pass both layers and 440 are excluded.

\section{Rubric Details}
\label{app:rubric}

This appendix complements the rubric composition (Equation~\ref{eq:rubric-composition}) and scoring (Equation~\ref{eq:rubric}) in Sections~\ref{sec:construction} and~\ref{sec:evaluation}.

\paragraph{Weights and constraints.} Table~\ref{tab:weights} lists the weights. A rubric contains three to six positive criteria, at most three penalty criteria, and at most eight criteria in total. When more than four criteria are mandatory, those with the lowest weights are downgraded to optional criteria with weight $1$ rather than removed.

\begin{table}[h]
\centering
\caption{Weights of rubric criteria.}
\label{tab:weights}
\small
\begin{tabular}{llr}
\toprule
Kind & Criterion & Weight \\
\midrule
Grounding & Core anchor of the instance & $+3$ \\
Grounding & Other anchors & $+2$ \\
Service   & Mandatory requirement of the request & $+2$ \\
Service   & Optional improvement & $+1$ \\
Penalty   & See Table~\ref{tab:penalty} & $-1$ to $-3$ \\
\bottomrule
\end{tabular}
\end{table}

\paragraph{Penalty criteria.} Table~\ref{tab:penalty} lists the five failure patterns summarised from negative user feedback, together with the condition under which each pattern is instantiated for an instance and the number of feedback items behind it. Every instantiated criterion cites the message that provides its evidence. A penalty criterion is phrased as a description of the failure and is judged as met when the response commits the failure, as instructed in the L2 evaluation prompt (Table~\ref{tab:prompt-l2-judge}).

\paragraph{Derivation of the patterns.} The five patterns come from negative feedback of our seed data, obtained in three steps.
\begin{enumerate}[leftmargin=*,itemsep=2pt]
  \item \emph{Collection.} Every member message that immediately follows an assistant reply and expresses a negative reaction (keyword filter): 423 candidates from five groups.
  \item \emph{Filtering.} A language model discards candidates that do not address the assistant, do not correct or object to the reply, or dispute subject knowledge rather than anything about the members: 97 remain (22.9\%).
  \item \emph{Clustering.} The 97 corrections are grouped by what the assistant got wrong: a member's current state (23), a recalled fact (24), an explicit agreement or persona (21), a superseded value (12), an earlier statement or settled plan (17).
\end{enumerate}

Four in five corrections concern the fidelity of the assistant's memory of members' states and facts rather than tone or knowledge, and the weights in Table~\ref{tab:penalty} follow this ranking; the superseded-value weight is raised to $-2$ for M3 instances, where that failure is the defining risk. The 97 items are recorded with the member's wording, e.g., ``not three, I did four sheets'' (recalled fact) and ``it should be geography now, not biology'' (superseded value), and are released with the benchmark.
The feedback is used once, to fix the set of patterns and their weights. Individual instances receive penalty criteria through the triggers of Table~\ref{tab:penalty}, which scan the anchors and the memory state of the instance; feedback from the instance's own group is not consulted.

\begin{table}[h]
\centering
\caption{Failure patterns of penalty criteria and their triggers.}
\label{tab:penalty}
\small
\begin{tabular}{p{0.28\linewidth}p{0.46\linewidth}rr}
\toprule
Failure pattern & Trigger & Feedback & Weight \\
\midrule
Distorting a recalled fact & An anchor contains a numeric or other exact value that is not already covered by a grounding criterion & 24 & $-2$ \\
Contradicting a settled plan & The group has settled a plan that is relevant to the request before the cutoff & 17 & $-1$ \\
Violating an explicit agreement or member constraint & A constraint is stated explicitly in the group, rather than inherited from a profile, and is relevant to the request & 21 & $-2$ \\
Misreading the current state of a member & A member states a current state within 150 messages before the cutoff & 23 & $-3$ \\
Relying on a superseded value & The instance contains a verified conflict pair & 12 & $-2$ \\
\bottomrule
\end{tabular}
\end{table}

\section{Human Validation}
\label{app:human}

We validated three things by hand: the instances themselves, the rubrics of the L2 instances, and the criterion-level judgments of the automatic judge. All three studies use the same sample of 120 instances drawn from the evaluation set, stratified by task level, memory dependency, and source (10 per cell), excluding instances that the pipeline had marked as superseded, and the same three annotators: one author and two annotators from the data provider, all fluent in Chinese and familiar with the group-chat setting. Annotators worked independently and did not see the outputs of the gates, the model reviewer, or the judge. An earlier annotation of the same sample by a second author was excluded: a joint review found that it had applied a stricter reading of the guidelines than the guidelines intended (it rejected 24\% of the instances, against 5--10\% for the three annotators, and agreed with them on only 71--79\% of the instances), and the guidelines were clarified before the annotation reported here. We report raw agreement rather than a chance-corrected coefficient: valid judgments dominate every annotator's marginals, so coefficients such as Cohen's $\kappa$ have almost no room above chance and are uninformative here.

\subsection{Instance validity}
\label{app:human-instances}

The annotators judged each instance on four questions: (Q1) whether each anchor is established by the cited message, (Q2) whether each anchor is necessary for a good response, (Q3) whether the reference facts (L1) or the rubric (L2) are valid, and (Q4) for M3 instances, whether the two values concern the same matter and the new value is current at the cutoff. Each judgment is binary; an instance is \emph{valid} if it passes all applicable questions. Table~\ref{tab:human-validation} reports the results.

\begin{table}[h]
\centering
\small
\caption{Human validation on 120 evaluation instances by three annotators. \emph{Majority}: at least two of three accept; \emph{unanimous}: all three accept; agreement is the pairwise agreement for the three annotator pairs. Q4 applies to the 40 M3 instances only. An instance is valid when every applicable question passes.}
\label{tab:human-validation}
\resizebox{\linewidth}{!}{%
\begin{tabular}{l r rrr rr r}
\toprule
 & & \multicolumn{3}{c}{Valid, per annotator (\%)} & \multicolumn{2}{c}{Valid (\%)} & \\
\cmidrule(lr){3-5}\cmidrule(lr){6-7}
Question & $n$ & 1 & 2 & 3 & Majority & Unanimous & Agreement (\%) \\
\midrule
Q1 Anchor established by cited message & 120 & 99.2 & 97.5 & 97.5 &  98.3 & 96.7 & 98.3 / 98.3 / 98.3 \\
Q2 Anchor necessary                    & 120 & 96.7 & 96.7 & 95.8 &  97.5 & 94.2 & 98.3 / 96.7 / 96.7 \\
Q3 Answer key or rubric valid          & 120 & 97.5 & 95.8 & 95.0 &  97.5 & 93.3 & 98.3 / 96.7 / 95.8 \\
Q4 Conflict pair valid (M3)            &  40 & 97.5 & 95.0 & 97.5 & 100.0 & 90.0 & 92.5 / 95.0 / 92.5 \\
\midrule
Instance valid (all questions)         & 120 & 95.0 & 91.7 & 90.0 &  95.8 & 84.2 & 93.3 / 91.7 / 90.0 \\
\bottomrule
\end{tabular}%
}
\end{table}

By majority vote, 95.8\% of the instances are valid (93.3\% L1, 98.3\% L2; 95.0/92.5/100\% for M1/M2/M3), and every question is accepted by at least 97.5\% by majority. Agreement is uniform across the three pairs (90--93\% on the instance level, 93--98\% per question), with no annotator systematically stricter than the others: of the 480 judgments, 464 are unanimous, 15 are rejected by a single annotator, and one is rejected by two. The unanimous rate (84.2\%) is lower than the majority rate because an instance must pass up to four independent checks. The five instances rejected by majority fail on an anchor whose stated fact over-interprets a tentative message (Q1, 2 instances) or on an anchor that is not needed for the response together with a reference answer or rubric that goes beyond the request (Q2 and Q3, 3 instances). We did not verify anchor completeness, i.e., whether a required memory is missing, because it requires reading the whole history before the cutoff; completeness relies on the two-pass extraction and the counterfactual check of Appendix~\ref{app:construction}. The annotation cards and all judgments are released with the benchmark.

\subsection{Rubric quality}
\label{app:human-rubric}

The instance-level check asks whether a rubric is valid as a whole. To examine the rubrics more closely, the annotators judged the 60 L2 instances of the sample on three further questions: (R1) whether every required service criterion is a reasonable, checkable requirement for the request rather than a restatement of it or a generic demand; (R2) whether every penalty criterion cites a fact relevant to the request and can be decided from the response; and (R3) whether a response that satisfies the rubric would be a good response in the situation, with no missing requirement, no contradiction, and no clearly misplaced weight. Majority-vote acceptance is 98.3\% for R1, 95.0\% for R2, and 96.7\% for R3; 90.0\% of the rubrics pass all three by majority and 86.7\% unanimously. Pairwise agreement is 0.93--1.00 per question and 0.90--0.97 on all three. Penalty criteria are the weakest component: the two rubrics rejected unanimously fail on R2 or R3, and M3 rubrics have the lowest pass rate (80.0\%, against 90.0\% and 100\% for M1 and M2), consistent with the fallback rule that derives a penalty from the superseded value when no user feedback is available (Appendix~\ref{app:rubric}).

\subsection{Agreement between the judge and human annotators}
\label{app:human-judge}

We then checked the criterion-level judgments of the automatic judge. For each of the 60 L2 instances, the annotators independently marked every criterion as met or not met for three real responses (full-context GPT-5.6-terra, dense retrieval, and EverMemOS), presented in random order under anonymous labels and without access to the judge's output, giving 384 criteria $\times$ 3 responses $=$ 1,152 judgments per annotator. Table~\ref{tab:judge-agreement} compares the judge with the majority vote of the annotators and with the annotators among themselves.

\begin{table}[h]
\centering
\small
\caption{Criterion-level agreement between the automatic judge (GPT-5.6-sol) and three human annotators on 60 L2 instances $\times$ 3 responses. Human--human is the mean pairwise agreement; judge--human compares the judge with the annotators' majority vote. Precision and recall are for the \emph{met} label with the majority vote as reference. Miss: the majority marks a criterion met and the judge does not; extra: the reverse. For penalty criteria, \emph{met} means the response committed the error.}
\label{tab:judge-agreement}
\resizebox{\linewidth}{!}{%
\begin{tabular}{l r rr rr rr}
\toprule
 & & \multicolumn{2}{c}{Agreement} & \multicolumn{2}{c}{Judge vs.\ majority} & \multicolumn{2}{c}{Disagreement (\%)} \\
\cmidrule(lr){3-4}\cmidrule(lr){5-6}\cmidrule(lr){7-8}
Criterion type & $n$ & Human--human & Judge--human & Precision & Recall & Miss & Extra \\
\midrule
Grounding (G)         &   351 & 0.85 & 0.86 & 1.00 & 0.67 & 14.2 & 0.0 \\
Service, required (S) &   300 & 0.88 & 0.89 & 0.99 & 0.80 & 10.7 & 0.3 \\
Service, bonus (S)    &   180 & 0.86 & 0.86 & 0.92 & 0.70 & 11.7 & 2.2 \\
Penalty (P)           &   321 & 0.99 & 0.95 & 0.48 & 1.00 &  0.0 & 4.7 \\
\midrule
All                   & 1,152 & 0.90 & 0.89 & 0.94 & 0.74 &  8.9 & 1.7 \\
\bottomrule
\end{tabular}%
}
\end{table}

The judge agrees with the human majority on 89.3\% of the judgments, the same level as the annotators agree with one another (mean pairwise 0.898, range 0.88--0.93), and it agrees with the strictest annotator (0.915) more often than that annotator agrees with the other two. Its errors are one-directional. On grounding and service criteria the judge only misses: it never credits a grounding criterion the majority rejects, but fails to credit 14.2\% of the grounding criteria and about 11\% of the service criteria that the majority accepts, typically when the response states the fact in paraphrase rather than in the wording of the rubric. On penalty criteria it only over-penalizes: it never overlooks an error the majority found, but flags 4.7\% of the penalty criteria that the majority considers not violated. Both directions lower the scores; neither raises them. At the instance level, recomputing score and pass from the human judgments with the formula of Section~\ref{sec:evaluation} gives a pass agreement of 88.9\% and a score correlation of $r=0.84$ ($\rho=0.84$) with the judge, above the mean between annotators (86.3\% and $r=0.81$); the judge's mean score is 0.13 below the human majority, and the ranking of the three systems is identical under the judge and under every annotator. Responses that decline to answer (25 of the 60 EverMemOS responses) are not a source of disagreement: on these the judge and the majority agree on 95--100\% of the criteria. We therefore read the automatic scores as a conservative estimate that preserves the ordering of systems, and in Section~\ref{sec:analysis} we treat the share of ``present but unused'' memories as an upper bound that the miss rate of the judge would reduce by roughly one seventh.

\section{Dataset Statistics}
\label{app:evaluation-scope}
\label{app:benchmark-composition}

\bench{} comprises 2,395 questions grounded in conversations from 50 groups: 1,374 questions for memory recovery (L1) and 1,021 for memory use (L2). Each question is assigned to one of three memory dependency levels. M1 (\emph{Single}) requires one substantive memory, M2 (\emph{Integrative}) combines complementary memories, and M3 (\emph{Evolving}) resolves updates, conflicts, or differences in applicability. The corpus contains 80,299 message records and 4,924,259 tokens. Table~\ref{tab:dataset-statistics} summarizes the question distribution and variation across groups. Group histories range from 2,887 to 972,839 tokens, with ten groups exceeding 100,000 tokens and four exceeding 200,000 tokens.

\begin{table}[htbp]
\centering
\caption{Statistics of \bench{} across 50 groups. The upper block shows question counts by evaluation level and memory dependency. The lower block reports corpus totals and the minimum, mean, and maximum per group.}
\label{tab:dataset-statistics}
\begin{tabular}{lrrrr}
\toprule
\multicolumn{5}{l}{\textit{Question distribution}} \\
\midrule
Level & M1 & M2 & M3 & Total \\
\midrule
L1 & 776 & 448 & 150 & 1,374 \\
L2 & 221 & 650 & 150 & 1,021 \\
Total & 997 & 1,098 & 300 & 2,395 \\
\midrule
\multicolumn{5}{l}{\textit{Corpus scale}} \\
\midrule
Statistic & Total & Min. & Mean & Max. \\
\midrule
Questions & 2,395 & 4 & 47.90 & 110 \\
Message records & 80,299 & 68 & 1,605.98 & 4,442 \\
History tokens & 4,924,259 & 2,887 & 98,485.18 & 972,839 \\
\bottomrule
\end{tabular}
\end{table}

History lengths are measured over the complete packaged conversation of each group using \texttt{o200k\_base} from \texttt{tiktoken} 0.12.0. Records are serialized as \texttt{\#turn [speaker] content} and joined by newlines. The packaged text retains at most 2,000 characters per message. These counts precede the question cutoff and model context budget, and exclude member profiles, evaluation queries, and system prompts. They describe corpus length, not the context tokens supplied to a model for an individual question. Message counts refer to stored records, including distinct records that share a turn index.

\section{Implementation Setup}
\label{app:experimental-details}

\subsection{History Cutoffs}
\label{app:history-cutoffs}

Every question, whether taken from the conversation or authored for the benchmark, is assigned a history cutoff $c_i$ to prevent answer leakage from later messages. Only conversation turns with index $t<c_i$ are eligible for context or memory construction. The turn at $c_i$ and all subsequent turns are excluded from the historical input. The evaluation query is supplied separately. The same temporal restriction applies to both native and authored questions.

\subsection{Retrieval and Memory Configuration}
\label{app:method-configurations}

We build on the official implementations where available and adapt their data interfaces and configurations to the group conversations in \bench{}. The adaptations handle conversation records, speaker identifiers, and question cutoffs. Table~\ref{tab:recorded-method-settings} summarizes how each method accesses the available history. 
\begin{table*}[t]
\centering
\caption{Overview of history access and memory retrieval in the evaluated systems.}
\label{tab:recorded-method-settings}
\small
\begin{tabularx}{\textwidth}{@{}lX@{}}
\toprule
Method & History and memory access \\
\midrule
Full Context & Supplies conversation history before the cutoff directly to the task agent, with context fitting when needed. \\
BM25 & Uses overlapping windows of 10 messages with a stride of 5 messages. Ranks these windows and eligible profile entries with BM25 and retrieves the top 20 units. \\
Dense retrieval & Uses the same overlapping windows of 10 messages with a stride of 5 messages and eligible profile entries. Retrieves the top 20 units using embedding similarity. \\
GraphRAG & Uses local search over entities and relations to retrieve graph evidence and associated source passages. \\
RAPTOR & Builds a hierarchy of messages and summaries and retrieves relevant representations across its levels. \\
Mem0 & Extracts and updates memory entries from the conversation, then retrieves entries relevant to the query. \\
LightMem & Compresses incoming messages, organizes them by topic, and retrieves relevant memories. \\
A-MEM & Organizes linked memory notes and retrieves relevant notes together with their neighbours. \\
MemoryOS & Combines recent messages, retrieved sessions, and stored knowledge to provide query context. \\
xMemory & Uses adaptive retrieval over themes, semantic memories, and episodes. \\
EverMemOS & Retrieves episode memories by combining embedding and BM25 rankings through reciprocal rank fusion. \\
\bottomrule
\end{tabularx}
\end{table*}

\subsection{Generation and Output Validation}
\label{app:generation}

The API compatibility configuration for GPT-5.6-terra and GPT-5.6-sol uses the supported default temperature of 1, either explicitly or by omitting the temperature parameter. This accommodates the model interface rather than imposing a temperature shared by all model families. Generation uses the adapters' output limits and retry procedures.

The evaluator returns structured JSON. Its parser checks that judgments cover all required atoms or rubric criteria. An incomplete judgment is treated as a generation failure rather than filling missing judgments with false values. The final comparison additionally requires a nonempty answer, a matching source cutoff, and a judgment from the designated evaluator. Section~\ref{app:prompt-design} presents the answer generation and evaluation prompts.

\section{Experiments}
\label{app:experiments}

\subsection{Cost and Token Efficiency}
\label{app:efficiency-analysis}

We separate memory construction from retrieval and answer generation. Each evaluation item contributes one query, and costs are aggregated across the full dataset. In Figure~\ref{fig:cost-break-even}, construction cost is amortised over additional queries, while the mean cost of each query remains fixed.

Table~\ref{tab:stage-token-costs} reports token consumption for construction, retrieval, answer generation, and judging across the runs used in the main comparison. Construction reports LLM tokens and embedding tokens separately, while retrieval reports their combined total. GPT-5.6-sol serves as the judge model. Judge tokens are included for reproducibility but excluded from the deployment curves because judging is not performed during deployment.

\begin{table*}[t]
\centering
\small
\caption{Token consumption by stage. All numeric entries are in millions of tokens (M). Construction reports LLM and embedding tokens separately, while retrieval combines both types.}
\label{tab:stage-token-costs}
\setlength{\tabcolsep}{4pt}
\begin{tabular}{@{}lrrrrrr@{}}
\toprule
Method & \shortstack{Build\\LLM} & \shortstack{Build\\embed.} & Retrieval & Answer & Judge & Total \\
\midrule
Full Context (Claude Sonnet 5) & 0.0 & 0.0 & 0.0 & 249.4 & 3.3 & 252.8 \\
Full Context (Terra) & 0.0 & 0.0 & 0.0 & 183.4 & 3.2 & 186.6 \\
Full Context (DeepSeek) & 0.0 & 0.0 & 0.0 & 156.0 & 3.4 & 159.4 \\
Full Context (Qwen) & 0.0 & 0.0 & 0.0 & 164.6 & 3.7 & 168.2 \\
\midrule
BM25 & 0.0 & 0.0 & 0.0 & 37.8 & 3.2 & 41.0 \\
Dense retrieval & 0.0 & 9.7 & 0.2 & 33.8 & 3.2 & 46.8 \\
GraphRAG & 41.9 & 13.1 & 69.2 & 65.8 & 3.2 & 193.1 \\
RAPTOR & 22.1 & 17.6 & 0.2 & 50.9 & 3.2 & 94.0 \\
\midrule
Mem0 & 94.8 & 11.9 & 0.0 & 6.1 & 2.9 & 115.7 \\
LightMem & 13.9 & 0.0 & 3.6 & 4.5 & 2.6 & 24.6 \\
A-MEM & 260.0 & 28.6 & 13.5 & 20.1 & 3.3 & 325.5 \\
MemoryOS & 121.6 & 10.0 & 0.2 & 19.4 & 3.6 & 154.7 \\
xMemory & 116.1 & 11.4 & 0.3 & 21.0 & 3.3 & 152.0 \\
EverMemOS & 272.9 & 3.2 & 0.2 & 73.9 & 3.4 & 353.5 \\
\bottomrule
\end{tabular}
\end{table*}

For the deployment comparison, total cost after $Q$ queries is the construction cost plus the mean cost of retrieval and generation for each query multiplied by $Q$. Table~\ref{tab:deployment-cost-details} reports the values underlying Figure~\ref{fig:cost-break-even}. GraphRAG combines its retrieval LLM allocation with one query embedding, while MemoryOS, xMemory, and EverMemOS use their recorded query embedding usage.

\begin{table*}[t]
\centering
\small
\caption{Deployment cost statistics relative to Full Context with GPT-5.6-terra. Construction cost is paid once; the cost of each query includes retrieval and answer generation but excludes judging. $N^*$ is the number of queries required to offset construction cost, and $C$ is the cumulative cost for all 2,395 queries in the dataset.}
\label{tab:deployment-cost-details}
\begin{tabular}{@{}lrrrr@{}}
\toprule
Method & \shortstack{Construction\\cost (USD)} & \shortstack{Cost/query\\(USD)} & $N^*$ & $C$ (USD) \\
\midrule
Full Context (Terra) & 0.0 & 0.1088 & --- & 260.4 \\
\midrule
BM25 & 0.0 & 0.0253 & 0 & 60.6 \\
Dense retrieval & 0.2 & 0.0229 & 3 & 55.0 \\
GraphRAG & 264.6 & 0.0906 & 14,574 & 481.5 \\
RAPTOR & 127.6 & 0.0321 & 1,665 & 204.5 \\
\midrule
Mem0 & 232.9 & 0.0062 & 2,270 & 247.7 \\
LightMem & 57.7 & 0.0050 & 556 & 69.6 \\
A-MEM & 1,314.7 & 0.0146 & 13,954 & 1,349.6 \\
MemoryOS & 568.5 & 0.0144 & 6,022 & 602.9 \\
xMemory & 402.6 & 0.0142 & 4,257 & 436.6 \\
EverMemOS & 706.3 & 0.0454 & 11,138 & 814.9 \\
\bottomrule
\end{tabular}
\end{table*}

The accounting reveals two distinct efficiency regimes. BM25, dense retrieval, LightMem, and RAPTOR recover their construction costs within the dataset, but their low costs do not yield the strongest task scores. Conversely, several structured memory systems improve L1 recovery while requiring more queries to recover their construction costs. Overall memory efficiency should therefore consider evidence utility and service quality together with construction costs and expenditure during queries, rather than equating shorter context with a better memory system.

\subsection{Intervention Analysis}
\label{app:intervention-analysis}

\paragraph{Re-slicing by generation model.}
All retrieval and memory systems generate with GPT-5.6-terra, so the full-context run of the same model is the matched comparison. Among the 497 paired memories, the fraction of memories that are in the model's input and are then used is 66\% for full context with GPT-5.6-terra, 61--62\% for BM25, Dense and EverMemOS, 50\% for xMemory, 43--46\% for A-MEM, MemoryOS and Mem0, and 24\% for LightMem. The gap between EverMemOS and full context in Figure~\ref{fig:recall-use} is thus partly a generation-model gap (Claude Sonnet 5 and DeepSeek-V4-flash use 72\% and 79\% of in-input memories); the remaining memory systems fall behind the matched full-context model even when the memory is present.

\paragraph{Re-slicing by request type: native versus authored.}
L2 requests are either written by group members (native) or authored by the pipeline; authored requests are checked not to reveal the memory, but they ask about the matter the memory bears on, whereas native requests are ordinary messages. The two types have almost the same distribution of lexical overlap with the memory (30\%/49\%/21\% vs.\ 35\%/48\%/17\% of memories at zero, low and higher overlap), and memories enter the input equally often (69\% vs.\ 67\% for memory systems). Yet memories in the input are used less often on native requests: 58\% vs.\ 71\% for full context, 38\% vs.\ 61\% for retrieval, and 30\% vs.\ 51\% for memory systems, and the gap persists within each overlap band (e.g., at zero overlap 56\% vs.\ 65\%, 32\% vs.\ 52\%, 19\% vs.\ 41\%). The cue that drives memory use is therefore not shared wording but whether the request is framed around the matter at all; questions in existing memory benchmarks are, and requests in real group chats are not.

\paragraph{Relational Evidence Refinement.}
RER operates between retrieval and answer generation. Given a query and its retrieved evidence, it identifies items that concern the same members, events, or decisions, organises them to expose complementarity, chronology, updates, and conflicts, and adds missing supporting items from the eligible candidate pool when necessary. It preserves the original history cutoff and source identifiers.

\paragraph{Assistance Constraint Binding.}
ACB interprets retrieved memories as constraints on the current assistance response. For each constraint, it records the affected subject, scope, content, status, applicability, source identifiers, and an exact supporting quotation. Code verifies the source and quotation, removes duplicate or inapplicable constraints, and supplies the validated constraint block as a separate input to answer generation. Thus, ACB changes how retrieved evidence governs the response without adding new memories.

\begin{figure}[h]
    \centering
    \includegraphics[width=0.88\linewidth]{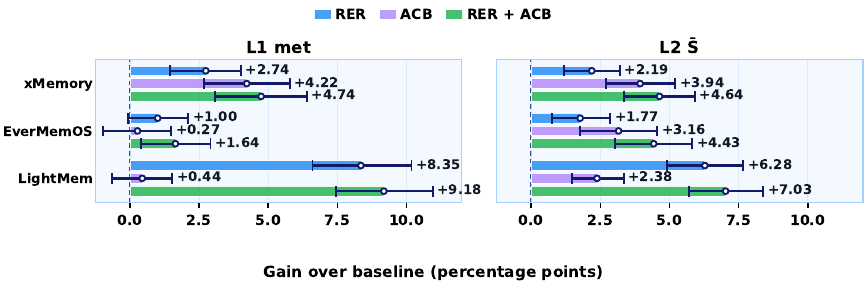}
    \caption{\textbf{Component analysis of RER and ACB.} Bars show gains over baseline in percentage points, with 95\% confidence intervals.}
    \label{fig:axis-combination}
\end{figure}

Table~\ref{tab:intervention-summary} provides the complete numerical results underlying Figure~\ref{fig:axis-combination}. RER changes the organisation or completeness of retrieved evidence, ACB binds retrieved evidence to constraints on the assistance response, and their combination applies both interventions. L1 changes are measured in Atom Recall points, while L2 changes are measured in rubric score points.

\begin{table*}[t]
\centering
\small
\caption{Performance gains from Relational Evidence Refinement (RER), Assistance Constraint Binding (ACB), and their combination. Values are percentage point gains over the corresponding baseline; brackets show the 95\% intervals used in Figure~\ref{fig:axis-combination}.}
\label{tab:intervention-summary}
\setlength{\tabcolsep}{5pt}
\begin{tabular}{@{}llcccc@{}}
\toprule
System & Level & RER & ACB & RER + ACB & Best condition \\
\midrule
\multirow{2}{*}{xMemory}
& L1 met & $2.74\ [1.45,4.01]$ & $4.22\ [2.66,5.77]$ & $4.74\ [3.08,6.40]$ &  RER + ACB \\
& L2 $\bar{S}$ & $2.19\ [1.18,3.21]$ & $3.94\ [2.70,5.21]$ & $4.64\ [3.36,5.92]$ & RER + ACB \\
\midrule
\multirow{2}{*}{EverMemOS}
& L1 met & $1.00\ [-0.09,2.11]$ & $0.27\ [-0.98,1.49]$ & $1.64\ [0.38,2.91]$ & RER + ACB \\
& L2 $\bar{S}$ & $1.77\ [0.74,2.85]$ & $3.16\ [1.77,4.55]$ & $4.43\ [3.02,5.82]$  & RER + ACB \\
\midrule
\multirow{2}{*}{LightMem}
& L1 met & $8.35\ [6.60,10.18]$ & $0.44\ [-0.65,1.53]$ & $9.18\ [7.46,10.95]$ & RER + ACB \\
& L2 $\bar{S}$ & $6.28\ [4.92,7.67]$ & $2.38\ [1.48,3.34]$ & $7.03\ [5.71,8.39]$ & RER + ACB \\
\bottomrule
\end{tabular}
\end{table*}

Table~\ref{tab:anchor-coverage-interventions} reports anchor coverage separately because it describes the evidence supplied to the task agent rather than answer quality. The table includes all four experimental conditions for the two hierarchical systems, whose memory units retain links to source turns. LightMem is omitted because its rewritten units combine and redistribute content from multiple turns, preventing unique coverage measurement at the turn level.

\begin{table*}[t]
\centering
\small
\caption{Anchor coverage under Baseline, RER, ACB, and RER + ACB for hierarchical external memory systems. Values report the fraction of required source turns retained in the evidence.}
\label{tab:anchor-coverage-interventions}
\setlength{\tabcolsep}{6pt}
\begin{tabular}{@{}lllccc@{}}
\toprule
System & Level & Condition & M1 & M2 & M3 \\
\midrule
\multirow{8}{*}{xMemory}
& \multirow{4}{*}{L1} & Baseline & 0.760 & 0.641 & 0.677 \\
& & RER & 0.760 & 0.641 & 0.677 \\
& & ACB & 0.760 & 0.639 & 0.674 \\
& & RER + ACB & 0.762 & 0.640 & 0.682 \\
\cmidrule(lr){2-6}
& \multirow{4}{*}{L2} & Baseline & 0.482 & 0.444 & 0.515 \\
& & RER & 0.482 & 0.444 & 0.515 \\
& & ACB & 0.477 & 0.445 & 0.515 \\
& & RER + ACB & 0.480 & 0.445 & 0.513 \\
\midrule
\multirow{8}{*}{EverMemOS}
& \multirow{4}{*}{L1} & Baseline & 0.858 & 0.796 & 0.849 \\
& & RER & 0.831 & 0.752 & 0.817 \\
& & ACB & 0.861 & 0.795 & 0.849 \\
& & RER + ACB & 0.830 & 0.750 & 0.815 \\
\cmidrule(lr){2-6}
& \multirow{4}{*}{L2} & Baseline & 0.586 & 0.645 & 0.681 \\
& & RER & 0.464 & 0.523 & 0.581 \\
& & ACB & 0.588 & 0.649 & 0.679 \\
& & RER + ACB & 0.456 & 0.516 & 0.576 \\
\bottomrule
\end{tabular}
\end{table*}

The component analysis exposes different system bottlenecks. RER contributes most to flat LightMem through evidence completion, whereas ACB contributes more to hierarchical xMemory and EverMemOS on L2 through response conditioning. Their combination performs best for every system on both levels, although the gains overlap rather than add directly. Figure~\ref{fig:axis-combination} shows the same pattern with confidence intervals; the two modules thus address the two bottlenecks of Figure~\ref{fig:recall-use} rather than the same one.

\section{Case Studies}
\label{sec:appendix_case}

\newtcolorbox{casebox}[2][]{
  enhanced, colback=white, colframe=black!55,
  boxrule=0.6pt, arc=3pt, left=6pt, right=6pt, top=4pt, bottom=4pt,
  fonttitle=\bfseries\small, coltitle=black, colbacktitle=black!8,
  attach boxed title to top left={yshift=-2pt,xshift=6pt},
  boxed title style={colframe=black!55,arc=2pt},
  title={#2}, #1
}

The following examples cover all six cells of the task taxonomy and are selected from the 2,395-item common evaluation set. Questions, memories, and responses are translated and paraphrased, and identifying details are replaced while preserving the evaluated relations. All responses use GPT-5.6-terra as the task agent and GPT-5.6-sol as the judge. For L1, the displayed score is Atom Recall; for L2, it is the normalized rubric score. In L2 rubrics, G denotes grounding, S denotes service quality, and P denotes a penalty.

\begin{casebox}{L1M1 (Single Memory): Identifying a loading failure}
\label{case:l1m1}
\footnotesize
\textbf{Question:} A mobile error trace was shared earlier. What failed, and at which loading step?\\[1pt]
\textbf{Memory anchor:}
\begin{enumerate}\setlength\itemsep{0pt}\setlength\topsep{1pt}
\item[$A_1$.] An Android application crashed with an \texttt{UnsatisfiedLinkError} while loading \texttt{libjiagu.so}.
\end{enumerate}
\textbf{Reference answer:} The application crashed because loading \texttt{libjiagu.so} raised an \texttt{UnsatisfiedLinkError}.

\begin{tcolorbox}[colback=green!3,colframe=green!45!black,boxrule=0.4pt,arc=1.5pt,left=4pt,right=4pt,top=2pt,bottom=2pt]
\textbf{Correct response (EverMemOS):} ``The Android application crashed with an \texttt{UnsatisfiedLinkError} while loading \texttt{libjiagu.so}.''\\[1pt]
\textit{Judge result:} Atom Recall $=1.00$ ($1/1$ reference atom recovered).
\end{tcolorbox}
\begin{tcolorbox}[colback=red!3,colframe=red!45!black,boxrule=0.4pt,arc=1.5pt,left=4pt,right=4pt,top=2pt,bottom=2pt]
\textbf{Incorrect response (LightMem):} The response attributes the trace to a different software issue.\\[1pt]
\textit{Judge result:} Atom Recall $=0.00$ ($0/1$); the failed library-loading step is not recovered.
\end{tcolorbox}
\end{casebox}

\medskip
\begin{casebox}{L1M2 (Multi-Memory Integration): Combining members' reactions}
\label{case:l1m2}
\footnotesize
\textbf{Question:} Three members reacted to the video just shared. Match each member to what they said, and list the three reactions.\\[1pt]
\textbf{Memory anchors:}
\begin{enumerate}\setlength\itemsep{0pt}\setlength\topsep{1pt}
\item[$A_1$.] Member A said the video left them stunned.
\item[$A_2$.] Member B found the sound painfully harsh.
\item[$A_3$.] Member C said they were also frightened.
\end{enumerate}
\textbf{Reference answer:} Member A was stunned by the video; Member B found its sound painfully harsh; Member C was also frightened.

\begin{tcolorbox}[colback=green!3,colframe=green!45!black,boxrule=0.4pt,arc=1.5pt,left=4pt,right=4pt,top=2pt,bottom=2pt]
\textbf{Correct response (xMemory):} ``A was stunned, B found the sound harsh, and C was also frightened.''\\[1pt]
\textit{Judge result:} Atom Recall $=1.00$ ($3/3$ reference atoms recovered with the correct speakers).
\end{tcolorbox}
\begin{tcolorbox}[colback=red!3,colframe=red!45!black,boxrule=0.4pt,arc=1.5pt,left=4pt,right=4pt,top=2pt,bottom=2pt]
\textbf{Incorrect response (LightMem):} The response assigns the reactions to different members and adds unrelated details.\\[1pt]
\textit{Judge result:} Atom Recall $=0.00$ ($0/3$ reference atoms recovered).
\end{tcolorbox}
\end{casebox}

\medskip
\begin{casebox}{L1M3 (Memory Updates and Conflicts): Resolving an updated itinerary}
\label{case:l1m3}
\footnotesize
\textbf{Question:} What is the current plan for tomorrow morning and noon, and what happened to the location that required a reservation?\\[1pt]
\textbf{Memory anchors:}
\begin{enumerate}\setlength\itemsep{0pt}\setlength\topsep{1pt}
\item[$A_1$.] The group will attend a banquet at noon.
\item[$A_2$.] Tickets for the originally planned location could not be reserved.
\item[$A_3$.] The morning plan was updated to visit an old-town area instead.
\end{enumerate}
\textbf{Reference answer:} Visit the old town in the morning, attend the banquet at noon, and note that the original location could not be booked. The updated morning plan supersedes the earlier itinerary.

\begin{tcolorbox}[colback=green!3,colframe=green!45!black,boxrule=0.4pt,arc=1.5pt,left=4pt,right=4pt,top=2pt,bottom=2pt]
\textbf{Correct response (EverMemOS):} ``Visit the old town in the morning and attend the banquet at noon; the original location could not be booked.''\\[1pt]
\textit{Judge result:} Atom Recall $=1.00$ ($3/3$ current reference atoms recovered).
\end{tcolorbox}
\begin{tcolorbox}[colback=red!3,colframe=red!45!black,boxrule=0.4pt,arc=1.5pt,left=4pt,right=4pt,top=2pt,bottom=2pt]
\textbf{Incorrect response (LightMem):} ``Follow the earlier sightseeing route and decide the remaining stops later.''\\[1pt]
\textit{Judge result:} Atom Recall $=0.00$ ($0/3$); the response returns unrelated itinerary fragments and misses the current facts.
\end{tcolorbox}
\end{casebox}

\medskip
\begin{casebox}{L2M1 (Single Memory): Drafting a group announcement}
\label{case:l2m1}
\footnotesize
\textbf{Request:} Turn the group's purpose and participation expectations into a concise announcement.\\[1pt]
\textbf{Memory anchor:}
\begin{enumerate}\setlength\itemsep{0pt}\setlength\topsep{1pt}
\item[$A_1$.] The group is for casual discussion of astrological charts, signs, and transits.
\end{enumerate}
\textbf{Rubric:}
\begin{enumerate}\setlength\itemsep{0pt}\setlength\topsep{1pt}
\item[G, +3.] State the remembered purpose of the group.
\item[S1, +2.] Use a concise, itemized announcement format.
\item[S2, +2.] Require friendly discussion, inclusion of newcomers, and tolerance of disagreement.
\item[S3, +2.] Encourage a rational attitude and prohibit fear-inducing or absolute claims.
\item[P1, $-2$.] Penalize any distortion of the remembered group purpose.
\end{enumerate}
The positive-score denominator is $9$.

\begin{tcolorbox}[colback=green!3,colframe=green!45!black,boxrule=0.4pt,arc=1.5pt,left=4pt,right=4pt,top=2pt,bottom=2pt]
\textbf{Correct response (LightMem):} ``Group notice: discuss charts, signs, and transits casually; welcome newcomers, respect disagreement, and avoid alarmist or absolute claims.''\\[1pt]
\textit{Rubric judge result:} G and S1--S3 are met; P1 is not triggered. Score: $(3+2+2+2)/9=1.00$.
\end{tcolorbox}
\begin{tcolorbox}[colback=red!3,colframe=red!45!black,boxrule=0.4pt,arc=1.5pt,left=4pt,right=4pt,top=2pt,bottom=2pt]
\textbf{Incorrect response (EverMemOS):} ``Please keep the group active and enjoy chatting with one another.''\\[1pt]
\textit{Rubric judge result:} G and S1--S3 are missed; P1 is not triggered. Score: $0/9=0.00$.
\end{tcolorbox}
\end{casebox}

\medskip
\begin{casebox}{L2M2 (Multi-Memory Integration): Coordinating field attendance}
\label{case:l2m2}
\footnotesize
\textbf{Request:} The discussion mixed who planned to attend an event, the transport arrangement, and who would remain behind. Summarize the current personnel arrangement in two parts.\\[1pt]
\textbf{Memory anchors:}
\begin{enumerate}\setlength\itemsep{0pt}\setlength\topsep{1pt}
\item[$A_1$.] Member A explicitly said they would attend the event on site.
\item[$A_2$.] The coordinator confirmed that five on-site members would travel together in one car.
\item[$A_3$.] Member B explicitly said they would remain with the rear-support team rather than attend on site.
\end{enumerate}
\textbf{Rubric:}
\begin{enumerate}\setlength\itemsep{0pt}\setlength\topsep{1pt}
\item[G1, +3.] State that Member B remains with rear support and is not in the on-site group.
\item[S1, +1.] Organize the answer into on-site attendance and rear support.
\item[G2, +2.] State that the five on-site members travel together in one car.
\item[S2, +2.] Preserve the single-car arrangement without inventing another roster or convoy.
\item[G3, +2.] Include Member A among those attending on site.
\item[P1, $-3$.] Penalize misattribution of another member's external-partnership status.
\item[P2, $-2$.] Penalize violation of the group's progress-reporting agreement.
\item[P3, $-1$.] Penalize contradiction of an existing contingency plan.
\end{enumerate}
The positive-score denominator is $10$.

\begin{tcolorbox}[colback=green!3,colframe=green!45!black,boxrule=0.4pt,arc=1.5pt,left=4pt,right=4pt,top=2pt,bottom=2pt]
\textbf{Correct response (EverMemOS):} ``On site: Member A joins the other attendees, and the five-person group travels together in one car. Rear support: Member B stays behind and does not join the on-site team.''\\[1pt]
\textit{Rubric judge result:} G1--G3 and S1--S2 are met; P1--P3 are not triggered. Score: $10/10=1.00$.
\end{tcolorbox}
\begin{tcolorbox}[colback=red!3,colframe=red!45!black,boxrule=0.4pt,arc=1.5pt,left=4pt,right=4pt,top=2pt,bottom=2pt]
\textbf{Incorrect response (LightMem):} ``Member A will attend on site. Member B can join the field team separately, and another vehicle can be arranged if needed.''\\[1pt]
\textit{Rubric judge result:} Only G3 is met; G1--G2 and S1--S2 are missed; P1--P3 are not triggered. Score: $2/10=0.20$.
\end{tcolorbox}
\end{casebox}

\medskip
\begin{casebox}{L2M3 (Memory Updates and Conflicts): Applying an updated production plan}
\label{case:l2m3}
\footnotesize
\textbf{Request:} Explain how an anniversary-video plan changed over time, identify who proposed or confirmed each change, and write a production brief that follows the final decision.\\[1pt]
\textbf{Memory anchors:}
\begin{enumerate}\setlength\itemsep{0pt}\setlength\topsep{1pt}
\item[$A_1$.] The initial proposal was a five-minute complete retrospective.
\item[$A_2$.] A member requested a maximum duration of three minutes.
\item[$A_3$.] Another member argued for a three-minute highlight edit to protect completion rate.
\item[$A_4$.] The final decision confirmed a three-minute montage synchronized to highlight moments.
\end{enumerate}
\textbf{Rubric:}
\begin{enumerate}\setlength\itemsep{0pt}\setlength\topsep{1pt}
\item[G1, +3.] Recover both the final three-minute duration and highlight synchronization.
\item[S1, +1.] Present the complete chronological update chain.
\item[G2, +2.] Attribute the duration limit and completion-rate rationale to the correct members.
\item[S2, +2.] Explain why the complete retrospective was replaced by a shorter montage.
\item[G3, +2.] Recover the initial five-minute proposal.
\item[P1, $-2$.] Penalize treating the superseded five-minute proposal as current.
\item[P2, $-2$.] Penalize violating the group's existing rule against public-shaming acknowledgments.
\item[P3, $-1$.] Penalize contradicting the agreed announcement time.
\end{enumerate}
The positive-score denominator is $10$.

\begin{tcolorbox}[colback=green!3,colframe=green!45!black,boxrule=0.4pt,arc=1.5pt,left=4pt,right=4pt,top=2pt,bottom=2pt]
\textbf{Correct response (EverMemOS):} ``The proposal moved from a five-minute retrospective to a three-minute maximum, then to a three-minute highlight montage to protect completion rate. Produce the final three-minute edit with cuts synchronized to the selected highlights.''\\[1pt]
\textit{Rubric judge result:} G1--G3 and S1--S2 are met; P1--P3 are not triggered. Score: $10/10=1.00$.
\end{tcolorbox}
\begin{tcolorbox}[colback=red!3,colframe=red!45!black,boxrule=0.4pt,arc=1.5pt,left=4pt,right=4pt,top=2pt,bottom=2pt]
\textbf{Incorrect response (BM25):} ``The team discussed an anniversary video, production pressure, delays, and a later editing handoff.''\\[1pt]
\textit{Rubric judge result:} G1--G3 and S1--S2 are missed; P1--P3 are not triggered. Score: $0/10=0.00$.
\end{tcolorbox}
\end{casebox}

\section{Prompting and Evaluation Protocol}
\label{app:prompt-design}

This section documents the task and evaluation protocol used for the reported experiments. We evaluate each method end to end using its official implementation where available, together with its method-specific memory construction and retrieval procedures. All retrieval and external memory systems use GPT-5.6-terra as the task agent, and all reported outputs are scored by GPT-5.6-sol with the same task-level criteria and scalar scoring rules. Method-specific context wrappers and response formats are retained where required by the corresponding implementations; these differences are treated as part of each complete system rather than as changes to benchmark questions or scoring.

\tcbset{benchmarkprompt/.style={
    width=\textwidth,
    colback=black!2,
    colframe=black!60,
    colbacktitle=black!10,
    coltitle=black,
    fonttitle=\bfseries,
    fontupper=\small,
    boxrule=0.5pt,
    arc=1mm,
    left=6pt, right=6pt, top=5pt, bottom=5pt,
    before upper={\raggedright}
}}

\subsection{Task and Evaluation Protocol}
\label{app:prompt-scope}

The source group conversations, runtime instructions, and evaluation prompts are in Chinese to preserve the original meaning and conversational context. We also provide a complete English translation of the benchmark for reference and future research. The boxes below give an English schematic specification of the shared task and evaluation requirements; they are not verbatim transcriptions of every adapter's wrapper. Braced placeholders denote inputs inserted at runtime, and character limits refer to Chinese characters.

Across methods, we hold fixed the benchmark query, history cutoff, language, task definition, and scoring criteria. The supplied context differs according to the history or memories made available by each method, while adapter-specific wrappers preserve the corresponding implementation interface. L1 outputs are scored against the same reference atoms and L2 outputs against the same instance-specific rubrics. Auxiliary diagnostics, where present, are excluded from the reported metrics.

\subsection{Answer Generation Prompts}
\label{app:answer-prompts}

Table~\ref{tab:prompt-unified-answer} summarizes the common answer requirements shared across implementations. Adapter-specific wrappers serialize full histories, retrieved passages, or stored memories according to the corresponding method interface, and the L1 response format is selected by task level.

\begin{table*}[!htbp]
\centering
\refstepcounter{table}\label{tab:prompt-unified-answer}
\begin{tcolorbox}[benchmarkprompt, title={Table~\thetable: Common Answer Requirements}]
\textbf{Instructions}\par\smallskip
You are an AI assistant in a group chat. Answer the question using the supplied context.
\begin{enumerate}
\setlength{\itemsep}{1pt}
\item Use only information supported by the context. Preserve the exact meaning of names, numbers, dates, and other factual details.
\item When records contain \texttt{\#index [speaker] content}, larger indices denote later turns. Refer to members using their supplied speaker identifiers.
\item If there are updates or conflicting claims, determine which information applies to the question, considering explicit corrections, decision authority, and the requested time or scope. Do not treat a superseded value as current.
\item If evidence is insufficient, explicitly state what cannot be determined from the context. Do not invent missing information or guess unsupported facts.
\item Answer in Chinese. Address the requested points concisely without repeating the question.
\end{enumerate}
\textbf{Response format}\par\smallskip
For L1, output only the following two fields:\par
\texttt{ANSWER:} Give the requested conclusions, with at most 20 Chinese characters per point. Separate multiple points with enumeration commas.\par
\texttt{WHY:} Briefly state the basis for the answer in at most 25 Chinese characters. Do not repeat \texttt{ANSWER}.\par\smallskip
For L2, answer the service request directly, following any format specified in the question and grounding the response in the supplied context.
\par\smallskip
\textbf{Input}\par\smallskip
Task level: \texttt{\{L1 or L2\}}\par
Context: \texttt{\{context\}}\par
Question: \texttt{\{query\}}
\end{tcolorbox}
\end{table*}

\subsection{Evaluation Prompts}
\label{app:judge-prompts}

Tables~\ref{tab:prompt-l1-judge} and~\ref{tab:prompt-l2-judge} summarize the common core of the L1 and L2 evaluation instructions. Reference atoms and rubric criteria are supplied only to the evaluator, not to the task agent. Every reported output is judged by GPT-5.6-sol under the same atom or rubric criteria, and auxiliary diagnostics do not enter the reported scores.

\begin{table*}[!htbp]
\centering
\refstepcounter{table}\label{tab:prompt-l1-judge}
\begin{tcolorbox}[benchmarkprompt, title={Table~\thetable: L1 Evaluation Prompt}]
You are a strict evaluator. Given a question, a list of atomic facts in the reference answer, and a candidate answer, independently determine whether the candidate recovers each atom.
\begin{enumerate}
\setlength{\itemsep}{1pt}
\item Each atom is either correct or incorrect. Allow paraphrases, but require equivalent numbers, names, dates, and times, such as 42.5 and 42.50, or 22:30 and 10:30 p.m.
\item If the answer is vague about an atom, asks a question instead, or says it does not know, mark that atom false.
\item Extra content outside the reference does not affect atom judgments. However, if the answer gives a value contradicting a reference atom, mark that atom false.
\item Separately count \texttt{extra}: concrete factual points that the reference did not request at all. Count each independent number, person, date, or factual matter. Do not count introductory wording, repetitions of the question, or explanatory filler. This count does not affect the atom judgments.
\end{enumerate}
Question: \texttt{\{query\}}\par
Atoms: \texttt{\{numbered reference atoms\}}\par
Answer: \texttt{\{candidate answer\}}\par\smallskip
Output only a JSON object with \texttt{atoms}, a Boolean array containing exactly \texttt{\{number of atoms\}} entries, \texttt{extra}, an integer, and \texttt{reason}, at most 25 Chinese characters.
\end{tcolorbox}
\end{table*}

For L2, each numbered criterion includes its kind, signed weight, and check text. The penalty marker is rendered as \texttt{Penalty} in the English specification.

\begin{table*}[!htbp]
\centering
\refstepcounter{table}\label{tab:prompt-l2-judge}
\begin{tcolorbox}[benchmarkprompt, title={Table~\thetable: L2 Evaluation Prompt}]
You are a strict evaluator. Independently judge whether the answer meets each criterion.
\begin{enumerate}
\setlength{\itemsep}{1pt}
\item \texttt{[G]} denotes memory grounding. The answer must explicitly state the specific fact referred to by the criterion, such as a number, name, date, chapter, or key point from an original statement. Appropriate behaviour without that specific fact is false.
\item \texttt{[q]} denotes a quality criterion. Judge according to its wording.
\item \texttt{[Penalty]} denotes a failure criterion. \texttt{met=true} means the answer actually commits the described error. A negatively phrased criterion such as ``must not misremember'' or ``must not contradict'' is a prohibition. For such a criterion, true still means violating the prohibition, not complying with it. When uncertain, return false.
\end{enumerate}
Question: \texttt{\{query\}}\par
Answer: \texttt{\{candidate answer\}}\par
Criteria: \texttt{\{numbered criteria with kind and signed weight\}}\par\smallskip
Output only a JSON array with exactly \texttt{\{number of criteria\}} entries. Each entry contains \texttt{n}, the criterion number, and \texttt{met}, a Boolean judgment.
\end{tcolorbox}
\end{table*}

The evaluation code computes the scalar scores from the returned judgments rather than asking the evaluator to assign a final score directly. A penalty marked true contributes its negative weight in Equation~\ref{eq:rubric}.

\section{Cost Accounting}
\label{app:cost-accounting}

We account for both computational and human costs throughout benchmark development. Computational costs include repeated data construction and validation, agent-assisted engineering, baseline adaptation and reruns, and final generation and evaluation. Recorded usage is priced according to the corresponding model, while incompletely logged adaptation and rerun costs are conservatively estimated from the main baseline runs. Human labor is grouped into benchmark design and protocol development, dataset construction and annotation, and quality assurance, adjudication, and experimental validation.

\begin{table}[h]
\centering
\small
\caption{Computational and human costs for benchmark development and baseline evaluation. We price benchmark construction, iteration, and agent-assisted engineering at GPT-5.5 rates (\$5 per million input tokens and \$30 per million output tokens). Baseline construction, generation, and judging use their recorded model prices, including GPT-5.6-sol for evaluation. Human labor costs reflect compensation actually paid to contributors. Paid workdays are converted to person-hours using an eight-hour workday, and person-hours, effective hourly rates, and USD values are rounded for presentation.}
\label{tab:cost-accounting}
\setlength{\tabcolsep}{4pt}
\resizebox{\linewidth}{!}{%
\begin{tabular}{@{}lrrr@{}}
\toprule
\multicolumn{4}{@{}l}{\textit{Computational costs}} \\
\midrule
Component & Input tokens & Output tokens & Cost (USD) \\
\midrule
Benchmark construction & 1{,}701.3M & 62.1M & 10{,}370 \\
Preliminary benchmark iteration & 315.8M & 8.4M & 1{,}832 \\
Agent-assisted benchmark engineering & 2{,}583.4M & 11.1M & 13{,}249 \\
\midrule
M-axis, L-axis, and combined analysis experiments & 706.5M & 31.2M & 1{,}871 \\
Baseline adaptation and reruns (estimated) & 961.1M & 103.9M & 3{,}076 \\
Main baseline evaluation & 1{,}922.3M & 207.8M & 6{,}151 \\
\midrule
\textbf{Computational subtotal} & \textbf{8{,}190.4M} & \textbf{424.6M} & \textbf{36{,}549} \\
\midrule
\multicolumn{4}{@{}l}{\textit{Human labor}} \\
\midrule
Component & Person-hours & Rate (USD/hour) & Cost (USD) \\
\midrule
Benchmark design and protocol development & 240 & 20 & 4{,}800 \\
Dataset construction and human annotation & 480 & 20 & 9{,}600 \\
Quality assurance, adjudication, and experimental validation & 160 & 15 & 2{,}400 \\
\midrule
\textbf{Human labor subtotal} & \textbf{880} & --- & \textbf{16{,}800} \\
\textbf{Total project cost} & --- & --- & \textbf{53{,}349} \\
\bottomrule
\end{tabular}%
}
\end{table}

The human-labor subtotal reports compensation actually paid to contributors. It covers benchmark and protocol design; construction and review of seed cases, queries, anchors, ground truths, and rubrics; M3 update and conflict annotation; L1--L2 pair verification; quality review and adjudication; and baseline adaptation, reruns, result verification, and statistical analysis. Converting paid workdays to eight-hour days, benchmark design involved two experts for approximately 15 days each (240 person-hours), dataset construction involved four experts for approximately 15 days each (480 person-hours), and quality assurance and validation involved ten experts for approximately two days each (160 person-hours). Automated model execution is excluded from person-hours; only active construction, review, debugging, monitoring, and validation time is counted.